\documentclass[10pt,twocolumn,letterpaper]{article}
\usepackage[pagenumbers]{iccv} 
\usepackage{pifont}
\definecolor{iccvblue}{rgb}{0.21,0.49,0.74}
\usepackage[pagebackref,breaklinks,colorlinks,allcolors=iccvblue]{hyperref}
\usepackage{multirow}
\usepackage{float}
\usepackage{makecell}
\usepackage{tabularx}
\newcolumntype{Y}{>{\raggedright\arraybackslash}X}

\def\paperID{3862} 
\def\confName{ICCV}
\def\confYear{2025}

\title{From Understanding to Erasing: Towards Complete and Stable Video Object Removal}

\author{
Dingming Liu$^{1}$ \quad
Wenjing Wang$^{2}$ \quad
Chen Li$^{2}$ \quad
Jing Lyu$^{2}$ \quad
Haotian Dong$^{2}$\\[3pt]
$^{1}$Peking University
\qquad
$^{2}$WeChat Vision, Tencent Inc.
}

\begin{document}
\twocolumn[{%
\renewcommand\twocolumn[1][]{#1}%
\maketitle

\begin{center}
    \centering
    \captionsetup{type=figure}
    \vspace{-5pt}
    \includegraphics[width=1.0\textwidth]{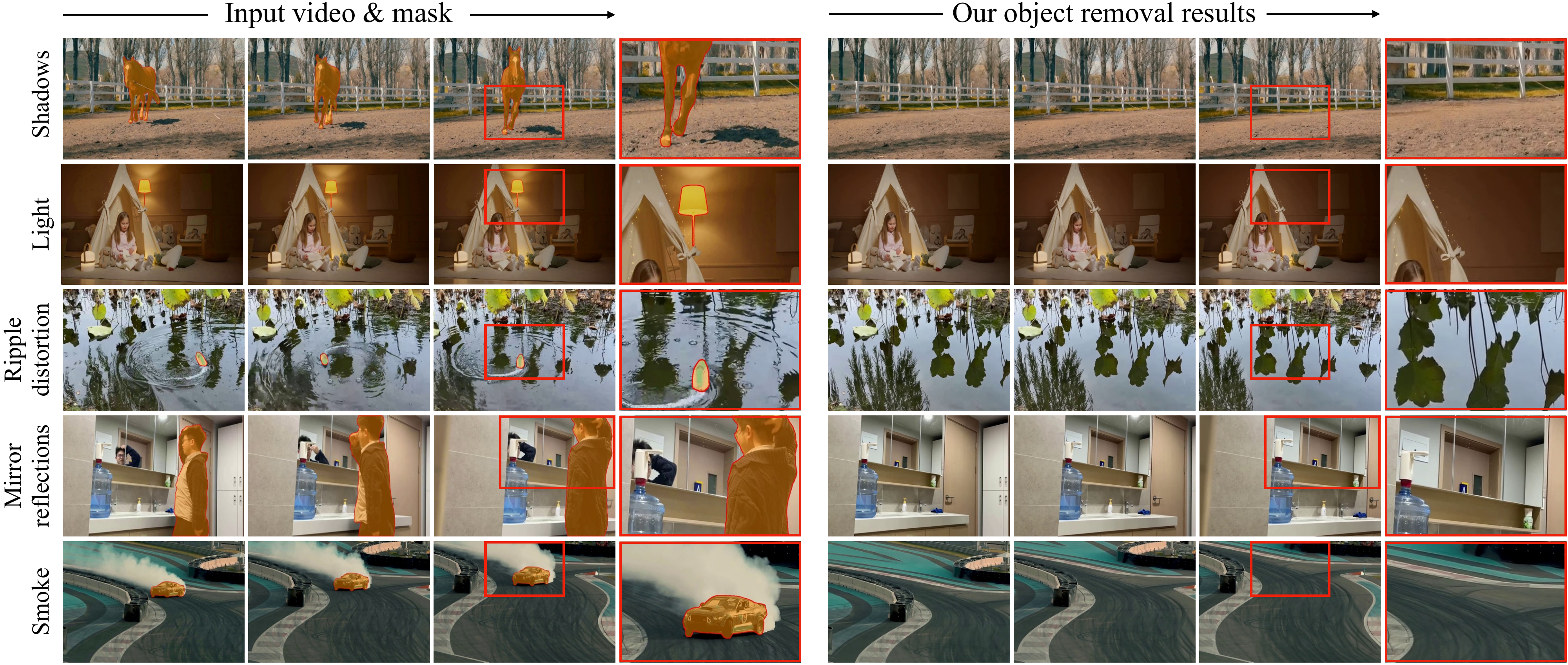}
    \captionof{figure}{
    Our method effectively removes target objects from videos together with diverse object-induced effects, including shadows, illumination variations, water-surface distortions, mirror reflections, and smoke.  Project page: \url{https://github.com/WeChatCV/UnderEraser}. 
    \label{fig:overview_results}}
\end{center}%
}]


\begin{abstract}
Video object removal aims to erase target objects while reconstructing visually plausible and temporally coherent content. However, target objects often induce shadows, reflections, illumination changes and other effects that extend beyond the provided mask, making conventional mask-conditioned completion prone to visible residuals. We therefore formulate side-effect-aware object removal as an understanding-guided process that integrates object--effect relations, affected-region localization, and context-aware reconstruction.
Specifically, we introduce Object-Induced Relation Distillation to transfer token-level object--effect relations from a pretrained vision foundation model to the video diffusion model.
We then design Object-aware Framewise Context Cross-Attention to combine target-object semantics with per-frame background context for removal and reconstruction, and propose Attention-guided Region Localization to derive a soft spatial prior over the target object and its affected regions. Extensive experiments across multiple benchmarks demonstrate that our method achieves more complete object-and-effect removal and outperforms existing approaches in removal quality, side-effect suppression, and temporal consistency.

\end{abstract}    
\section{Introduction}

Video object removal has emerged as a key visual editing technique~\cite{miao2025rose,fu2026effecterase} for erasing specified unwanted objects from videos and reconstructing the missing regions with contextually plausible content, while maintaining realistic visual quality and temporal consistency.

Early methods rely on 3D convolutions~\cite{chang2019free,hu2020proposal}, optical flow~\cite{xu2019deep,gao2020flow,kang2022error}, or Transformers~\cite{zeng2020learning,liu2021fuseformer,li2022towards} to propagate valid information into missing regions. Despite steady progress, they remain limited by local receptive fields, inaccurate motion estimation, or insufficient contextual support.

Recent diffusion-based methods~\cite{li2025diffueraser,zi2025minimax,miao2025rose} provide strong generative priors, but mostly treat object removal as mask-conditioned completion and thus overlook effects beyond the input mask. Although ROSE~\cite{miao2025rose} and EffectErase~\cite{fu2026effecterase} explicitly address object-induced effects, ROSE relies heavily on synthetic paired data and explicit difference-mask prediction, while EffectErase requires a relatively complex removal--insertion dual-task framework.

We argue that side-effect-aware object removal should be treated as an understanding problem: the model must recognize which visual effects are associated with the target object, locate them, and use the surrounding context to reconstruct the affected content. Based on this view, we formulate the task as an \emph{understanding-guided process} and propose an understanding-centric video diffusion framework.

To equip the model with object--effect understanding, we introduce Object-Induced Relation Distillation. We observe that pretrained vision foundation models provide more localized and fine-grained object-centric representations than those explicitly available in video diffusion models. Our method therefore transfers token-level object--effect relations from a vision foundation model to the video diffusion model, enabling it to capture associations between the target object and its induced effects.

Relational understanding alone, however, is insufficient to determine how the affected content should be reconstructed in each frame. We therefore introduce Object-aware Framewise Context Cross-Attention to ground object--effect understanding in frame-specific context. Target-object semantics indicate what should be removed, while background context provides evidence for what should be reconstructed, jointly guiding removal and completion. Moreover, object--effect understanding must be translated into spatial perception. We therefore introduce Attention-guided Region Localization, which derives a soft spatial prior from target-related feature responses and enhances the model's awareness of both the target object and its affected regions.

Together, these components constitute a progressive understanding-to-erasing process:
relation distillation identifies which visual effects are associated with the target object, contextual conditioning determines how the affected content should be reconstructed in each frame, and region localization enhances the model's spatial awareness of both the target object and its induced effects. Through this process, the model removes both the target object and its induced effects while preserving background fidelity and spatio-temporal consistency, as shown in Figure~\ref{fig:overview_results}. Extensive experiments on multiple benchmarks demonstrate that our method outperforms state-of-the-art video object removal approaches in removal quality, side-effect suppression, and temporal consistency, while generalizing well to challenging real-world scenarios.
Our contributions are summarized as follows:

\begin{itemize}

\item We formulate side-effect-aware video object removal as an understanding-guided process and propose an understanding-centric video diffusion framework. By transforming object--effect understanding into contextual and spatial guidance, our framework enables complete removal of both target objects and their associated effects.

\item We introduce Object-Induced Relation Distillation to transfer token-level object--effect relational knowledge from a pretrained vision foundation model to the video diffusion model, improving its understanding of target-induced effects.

\item We design Object-aware Framewise Context Cross-Attention to provide frame-level cues for removal and reconstruction, together with Attention-guided Region Localization to enhance the model's spatial awareness of the target object and its associated effects.

\end{itemize}
\section{Related Work}

\noindent \textbf{Video Diffusion Models.}
Recent diffusion-based approaches have made rapid progress in generating videos conditioned on text or images~\cite{chen2024videocrafter2,blattmann2023align,blattmann2023stable,ho2022imagen,guo2023animatediff,singer2022make}. AnimateDiff~\cite{guo2023animatediff} adapts pretrained text-to-image diffusion models for video synthesis by introducing dedicated motion components. Imagen Video~\cite{ho2022imagen} and Make-a-Video~\cite{singer2022make} learn spatiotemporal generation more directly by training cascaded spatial and temporal modules in pixel space.
These works typically use a U-Net~\cite{UNet} backbone.

Diffusion Transformer (DiT)~\cite{peebles2023scalable} has become the prevailing paradigm for video generation.
Representative systems such as Sora~\cite{brooks2024video}, CogVideoX~\cite{yang2024cogvideox}, HunyuanVideo~\cite{kong2024hunyuanvideo}, and StepVideo~\cite{ma2025step} combine DiT with large-scale training and improved latent video compression to enhance temporal coherence and synthesis quality.
Wan~\cite{wan2025wan} further advances this line of work by introducing a scalable spatiotemporal VAE. These models illustrate the growing effectiveness and versatility of DiT-based frameworks for high-resolution video synthesis.


\vspace{1mm}
\noindent \textbf{Video Object Removal.}
Video object removal aims to eliminate target objects while generating spatially and temporally coherent content.
Early works employed 3D CNNs ~\cite{chang2019free,hu2020proposal,wang2019video} to capture spatiotemporal features but were limited by small receptive fields and frame misalignment, reducing their ability to utilize distant context. Flow-based pixel propagation approaches~\cite{zhang2019internal,xu2019deep,gao2020flow,zhang2022inertia,kang2022error} effectively recover fine textures and details by leveraging neighboring frames. Transformer-based methods ~\cite{lee2019copy,li2020short,liu2021fuseformer,zeng2020learning,zhang2022flow,li2022towards,zhou2023propainter} leverage spatiotemporal attention to capture recurrent textures and contextual information across frames, enabling more effective reconstruction of object structures and coherent content in missing regions.

Recently, diffusion-based methods~\cite{wu2024towards,zhang2024avid,zi2025cococo,kushwaha2026object,lee2025video,li2025diffueraser,bian2025videopainter,miao2025rose,zi2025minimax,jiang2025vace,samuel2025omnimattezero,lee2025generative,fu2026effecterase,wu2026yose,chen2026generaser,chen2026draft} have become mainstream due to their superior ability to capture complex data distributions and their more stable performance. Among these, DiffuEraser~\cite{li2025diffueraser} adapts image inpainting model BrushNet~\cite{ju2024brushnet} for video inpainting through a two-stage training strategy. MiniMax-Remover~\cite{zi2025minimax} introduces a two-stage diffusion-based video object removal framework that eliminates textual conditioning and employs minimax distillation to achieve high-quality removal. ROSE~\cite{miao2025rose}, built on Wan2.1~\cite{wan2025wan}, introduces an auxiliary difference-mask predictor to explicitly localize regions modified by object removal. EffectErase~\cite{fu2026effecterase} introduces effect-aware video object removal by jointly learning removal and insertion. 

Despite these advances, existing methods typically rely on explicit affected-region prediction or task-specific training designs, while object--effect understanding and contextual reconstruction remain insufficiently integrated. 
\section{Methodology}

\begin{figure*}
\centering
\includegraphics[width=0.99\linewidth]{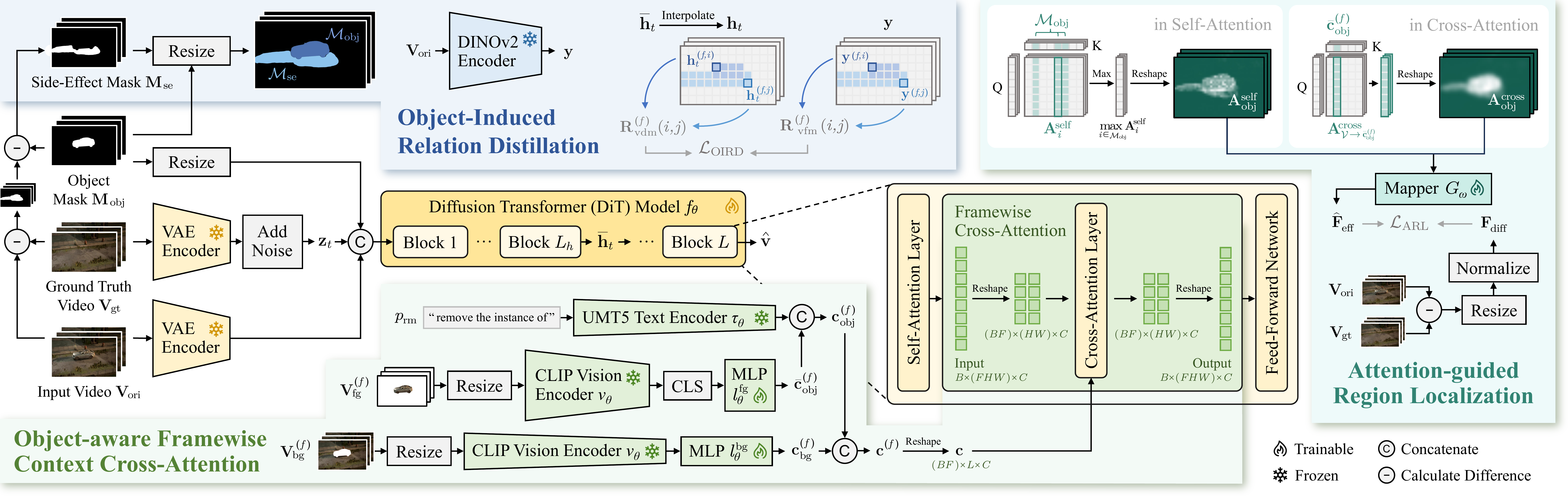}
\caption{Overview of our video object removal framework.
Given the input video, object mask, and object-removed ground truth, we construct the diffusion input by concatenating the noisy latent, encoded input video, and resized mask, and derive a side-effect mask from the difference between the input and ground-truth videos.
Object-Induced Relation Distillation transfers token-level object--effect relations from DINOv2 to the video diffusion transformer.
Object-aware Framewise Context Cross-Attention combines target-object semantics with per-frame background context to guide removal and reconstruction.
Attention-guided Region Localization fuses self- and cross-attention responses to predict a soft spatial prior over the target object and its affected regions, which is aligned with a difference-based affected-region cue.
}
\label{fig:framework}
\end{figure*}

\subsection{Overview}

Given an original video $\mathbf{V}_{\mathrm{ori}}$ and an object mask
$\mathbf{M}_{\mathrm{obj}}$, our goal is to remove both the target object
and its induced effects while preserving background fidelity and temporal
consistency. As shown in Figure~\ref{fig:framework}, our framework contains three
components. Object-Induced Relation Distillation transfers token-level
object--effect relations from DINOv2 to the video diffusion model.
Object-aware Framewise Context Cross-Attention combines target-object
semantics with frame-specific background context for removal and
reconstruction.
Attention-guided Region Localization extracts
object-related self- and cross-attention responses to localize affected
regions beyond the input mask.
The model is jointly optimized with the
standard flow-matching objective~\cite{lipman2022flow} and the proposed auxiliary losses.

\subsection{Object-Induced Relation Distillation}

\begin{figure}[t]
\centering
\includegraphics[width=0.99\columnwidth]{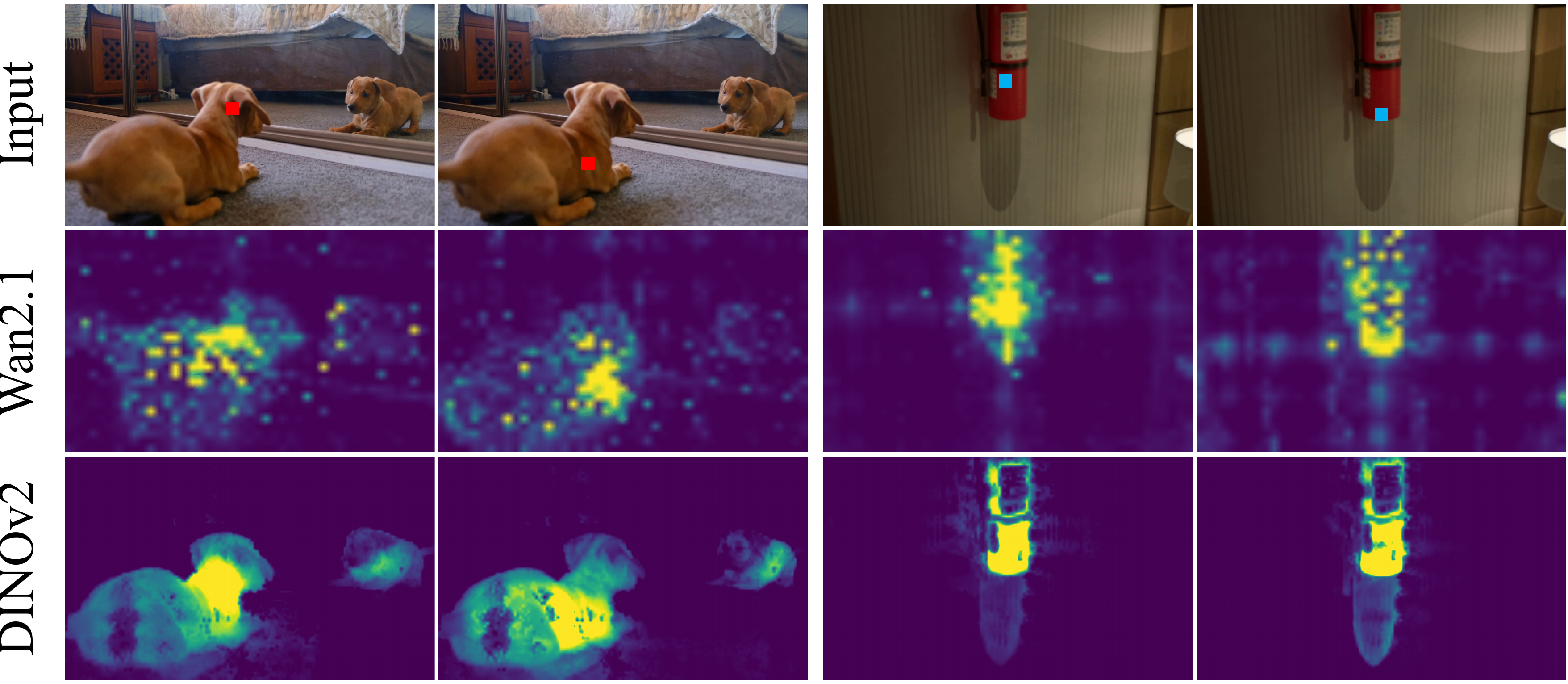}
\caption{
Comparison of the attention weights from all query tokens to selected key tokens within target object regions in Wan2.1 and DINOv2 for the same video frames. Wan2.1 exhibits incomplete and noisy responses to the target object and its induced side effects, while DINOv2 produces more accurate and localized token-wise responses, indicating stronger object--effect relational cues.
}
\label{fig:knowledge_com}
\end{figure}

\noindent\textbf{Motivation.}
Although pretrained diffusion models have shown certain visual perception capabilities~\cite{iccv_ZhaoRLLZL23_diff_for_perception,abs-2505-06890}, and recent studies suggest that video diffusion models may encode knowledge about object-induced side effects~\cite{lee2025generative,samuel2025omnimattezero}, we observe that such knowledge is not reliably organized at a fine-grained token level.
Figure~\ref{fig:knowledge_com} compares self-attention weights
from all query tokens to selected object-region key tokens
in our Video Diffusion Model (VDM) backbone, Wan2.1~\cite{wan2025wan}, and the Video Foundation Model (VFM),
DINOv2~\cite{oquab2023dinov2}. These attention maps reveal the object-centric token relations
captured by the two models.
Wan2.1 exhibits noisy and poorly localized token-wise self-attention, often missing object-induced effects or responding to irrelevant regions.
In contrast, DINOv2 produces more localized self-attention responses,
with different object tokens capturing distinct semantic parts and thus providing finer-grained object--effect relational cues.

This discrepancy reflects the different training objectives of VFMs and VDMs. VFMs are optimized for visual understanding and thus tend to learn object-centric relational representations, whereas VDMs are primarily optimized for generation and may not explicitly organize object--effect relations at the token level. We therefore distill object-induced relational knowledge from the VFM into the VDM to improve side-effect-aware video object removal.

\vspace{1mm}
\noindent\textbf{Distillation Design.}
The key idea is to align the object--effect relation of the VDM with that of a VFM. To this end, we first identify object and side-effect regions at the pixel level, then compare their token-level relational structures in the VDM and VFM feature spaces.

During training, given the input video $\mathbf{V}_{\mathrm{ori}}$, the corresponding object-removed ground-truth video $\mathbf{V}_{\mathrm{gt}}$ and the object mask $\mathbf{M}_{\mathrm{obj}}$, we derive a side-effect mask $\mathbf{M}_{\mathrm{se}}$ that localizes object-induced residual changes beyond the object region.
Following ROSE~\cite{miao2025rose}, we first compute a difference mask $\mathbf{M}_{\mathrm{diff}}$ by thresholding the per-pixel $\ell_2$ distance across channels:
\begin{equation}
\mathbf{M}_{\mathrm{diff}}(f,i,j)=
\mathbb{I}\!\left(
\left\|
\mathbf{V}_{\mathrm{ori}}(:,f,i,j) - \mathbf{V}_{\mathrm{gt}}(:,f,i,j)
\right\|_2
> \delta
\right),
\label{eq:mdiff}
\end{equation}
where $\mathbb{I}(\cdot)$ denotes the indicator function, $f$ indexes the frame, $(i,j)$ denotes the spatial pixel location, and $\delta>0$ is a fixed threshold ($0.1$ in our setting).
The resulting binary mask highlights pixel-level changes caused by object removal, including both the object region and its induced side effects.
Finally, we obtain the side-effect mask by excluding the object region:
\begin{equation}
\mathbf{M}_{\mathrm{se}} = \mathbf{M}_{\mathrm{diff}} \wedge \neg \ \mathbf{M}_{\mathrm{obj}},
\label{eq:mse}
\end{equation}
where $\wedge$ and $\neg$ denote logical AND and NOT, respectively.

We next extract the corresponding VDM and VFM features. During training, the ground truth video $\mathbf{V}_{\mathrm{gt}}$ is encoded by a Variational Autoencoder (VAE)~\cite{VAE} into the latent $\mathbf{z}_0$. We denote the corresponding noisy latent at continuous time $t \in [0,1]$ under the flow-matching parameterization as $\mathbf{z}_t$, defined as $\mathbf{z}_t = t \cdot \boldsymbol{\epsilon} + (1 - t) \cdot \mathbf{z}_0$, where $\boldsymbol{\epsilon}$ represents Gaussian noise.
Using $\mathbf{z}_t$, the resized object mask, and the encoded input video,

the DiT
$f_{\theta}$ produces an intermediate hidden representation $\bar{\mathbf{h}}_t$, which serves as the VDM spatiotemporal feature map. In parallel, we extract semantic features from the input video $\mathbf{V}_{\mathrm{ori}}$ using a frozen pretrained DINOv2 encoder, yielding $\mathbf{y} = \mathrm{DINO}(\mathbf{V}_{\mathrm{ori}})$, which captures semantic relations between the target object and its induced side effects.

We spatially interpolate each frame of
$\bar{\mathbf{h}}_t$ to the resolution of the DINOv2 feature grid,
yielding $\mathbf{h}_t$, such that $\mathbf{h}_t$ and $\mathbf{y}$
contain the same number of spatial tokens per frame.

Given these features, we first compute the token pairwise similarity matrix. The spatial relation matrix at frame $f$ is defined as:
\begin{equation}
\begin{aligned}
\mathbf{R}^{(f)}_{\mathrm{vdm}}(i,j)
&=\operatorname{sim}\!\left(
  \mathbf{h}_{t}^{(f,i)},\mathbf{h}_{t}^{(f,j)}
\right), \\
\mathbf{R}^{(f)}_{\mathrm{vfm}}(i,j)
&=\operatorname{sim}\!\left(
  \mathbf{y}^{(f,i)},\mathbf{y}^{(f,j)}
\right),
\end{aligned}
\label{eq:spatial_relation}
\end{equation}
where
$\operatorname{sim}(\mathbf{a},\mathbf{b})
=
\frac{\mathbf{a}^{\top}\mathbf{b}}
{\lVert\mathbf{a}\rVert_2\lVert\mathbf{b}\rVert_2}$,
$i,j\in\{1,\dots,N\}$ index spatial token positions,
$N$ denotes the number of spatial tokens in each frame (i.e., the spatial grid is flattened into a sequence of $N$ tokens), and
$\mathbf{R}^{(f)}_{\mathrm{vdm}},\mathbf{R}^{(f)}_{\mathrm{vfm}}
\in\mathbb{R}^{N\times N}$ denote the spatial relation matrices for frame $f$.

Let $\mathcal{M}_{\mathrm{obj}}^{(f)}$ and
$\mathcal{M}_{\mathrm{se}}^{(f)}$ denote the object-token and
side-effect-token index sets at frame $f$, obtained by resizing
$\mathbf{M}_{\mathrm{obj}}$ and $\mathbf{M}_{\mathrm{se}}$
to the aligned feature-grid resolution. Since some frames may contain no valid
side-effect tokens after thresholding and downsampling, we define the valid-frame set as
$\mathcal{F}_{\mathrm{valid}}
=
\left\{
f \;\middle|\;
|\mathcal{M}_{\mathrm{obj}}^{(f)}|>0,\;
|\mathcal{M}_{\mathrm{se}}^{(f)}|>0
\right\}$.

Finally, the proposed Object-Induced Relation Distillation (OIRD) loss is formulated as:

\begin{equation}
\mathcal{L}_{\mathrm{OIRD}} = 
\mathbb{E}_{f,i,j} \left[
\left|
\mathbf{R}_{\mathrm{vdm}}^{(f)}(i,j)
-
\mathbf{R}_{\mathrm{vfm}}^{(f)}(i,j)
\right|\right].
\end{equation}
Here, $f \sim \mathcal{F}_{\mathrm{valid}}$ indexes valid frames, while $i \sim \mathcal{M}_{\mathrm{obj}}^{(f)}$ and $j \sim \mathcal{M}_{\mathrm{se}}^{(f)}$ index object and
side-effect tokens, respectively. The expectations denote uniform
averaging over $\mathcal{F}_{\mathrm{valid}}$,
$\mathcal{M}_{\mathrm{obj}}^{(f)}$, and
$\mathcal{M}_{\mathrm{se}}^{(f)}$.
For training samples with $\mathcal{F}_{\mathrm{valid}}=\varnothing$, we set $\mathcal{L}_{\mathrm{OIRD}}=0$, while the remaining training objectives are computed as usual.

\subsection{Object-aware Framewise Context Cross-Attention}

After learning object--effect relations through OIRD, the DiT still requires frame-specific cues for removal and reconstruction. 

We therefore leverage the backbone’s native cross-attention blocks with two complementary conditions: an object-aware removal prompt that identifies the target instance and framewise background visual tokens that provide reconstruction context. Unlike prior methods that either rely on empty prompts~\cite{miao2025rose} or remove cross-attention blocks entirely~\cite{zi2025minimax}, our design preserves and explicitly exploits the backbone’s native cross-attention mechanism for understanding the removal foreground and background.

For each frame, we obtain the foreground object and the complementary background region as:
\begin{equation}
\mathbf{V}_{\mathrm{fg}} = \mathbf{V}_{\mathrm{ori}} \odot \mathbf{M}_{\mathrm{obj}},
\quad
\mathbf{V}_{\mathrm{bg}} = \mathbf{V}_{\mathrm{ori}} \odot (1-\mathbf{M}_{\mathrm{obj}}).
\end{equation}
The foreground object is used to construct an object-aware removal condition. Specifically, we encode the foreground object with the CLIP vision encoder and project its CLS token into the text-token embedding space through a lightweight MLP. We then concatenate this projected object token with the text embedding of the prompt ``remove the instance of'', forming the object-aware removal condition:
\begin{equation}
\bar{\mathbf{c}}_{\mathrm{obj}}^{(f)}
=
l_{\theta}^{\mathrm{fg}}
\left(
\mathrm{CLS}
\left(
\nu_{\theta}(\mathbf{V}_{\mathrm{fg}}^{(f)})
\right)
\right),
\ 
\mathbf{c}_{\mathrm{obj}}^{(f)}
=
\left[
\tau_{\theta}(p_{\mathrm{rm}}),
\;
\bar{\mathbf{c}}_{\mathrm{obj}}^{(f)}
\right],
\label{eq:condition_token_obj}
\end{equation}
where $p_{\mathrm{rm}}$ denotes the text prompt, $\tau_{\theta}(\cdot)$ and $\nu_{\theta}(\cdot)$ denote the frozen UMT5 text and CLIP vision encoders, respectively, $l_{\theta}^{\mathrm{fg}}(\cdot)$ is a trainable projection MLP, and $[\cdot, \cdot]$ denotes concatenation along the sequence dimension.

In parallel, the unmasked background region provides frame-specific context for reconstruction. We encode the background region of each frame using the CLIP vision encoder and project the resulting visual tokens into the cross-attention space:
\begin{equation}
\mathbf{c}_{\mathrm{bg}}^{(f)}
=
l_{\theta}^{\mathrm{bg}}
\left(
\nu_{\theta}(\mathbf{V}_{\mathrm{bg}}^{(f)})
\right).
\end{equation}

Because the background region may contain object-induced effects beyond the mask,  $l_{\theta}^{\mathrm{bg}}(\cdot)$ maps the framewise visual
tokens into a task-adapted conditioning space for reconstruction. The final framewise condition is:
\begin{equation}
\mathbf{c}^{(f)}
=
\left[
\mathbf{c}_{\mathrm{obj}}^{(f)},
\mathbf{c}_{\mathrm{bg}}^{(f)}
\right].
\end{equation}

Since standard video diffusion models typically use a single condition for the whole video, we enable framewise conditioning by merging the batch and frame dimensions. Given condition tokens with shape $[B \times F \times L \times C]$, we reshape them into:
\begin{equation}
\mathbf{c}
\in
\mathbb{R}^{(B F) \times L \times C}.
\end{equation}

The video latent tokens are reshaped in the same way before cross-attention and restored afterward, which is illustrated in Figure~\ref{fig:framework}. In this way, each frame receives dedicated
object-aware and background-aware conditioning, enabling
the model to exploit more detailed, frame-specific guidance
for accurate target removal and context-consistent
reconstruction.

\subsection{Attention-guided Region Localization}

While OIRD aligns feature-level object--effect relations and Object-aware Framewise Context Cross-Attention provides contextual guidance for removal and reconstruction, affected regions remain implicit in the learned representations.

We therefore design Attention-guided Region Localization to guide the DiT's object-related attention toward both the target object and its induced side effects.

For each frame (we omit the frame index in the following for simplicity), we
resize $\mathbf{M}_{\mathrm{obj}}$ to the spatial
resolution of the self-attention map and reuse the notation
$\mathcal{M}_{\mathrm{obj}}$ for the resulting object-token index set.
For each $i \in \mathcal{M}_{\mathrm{obj}}$, we extract its self-attention map $\mathbf{A}^{\mathrm{self}}_i$ from the DiT.
These self-attention maps describe how object tokens interact with other latent tokens in the video. To preserve the strongest object-related responses, we aggregate them using max pooling:
\begin{equation}
\mathbf{A}^{\mathrm{self}}_{\mathrm{obj}}
=
\max_{i \in \mathcal{M}_{\mathrm{obj}}}
\mathbf{A}^{\mathrm{self}}_i .
\end{equation}
Meanwhile, we reuse the object-aware removal condition $\bar{\mathbf{c}}_{\mathrm{obj}}^{(f)}$ defined in Eq.(\ref{eq:condition_token_obj}) and extract the cross-attention map corresponding to the projected object token:
\begin{equation}
\mathbf{A}^{\mathrm{cross}}_{\mathrm{obj}}
=
\mathbf{A}^{\mathrm{cross}}_{\mathcal{V} \rightarrow \bar{\mathbf{c}}_{\mathrm{obj}}^{(f)}},
\end{equation}
where $\mathcal{V}$ denotes all video latent tokens.

The self- and cross-attention maps provide complementary localization cues from latent object--context dependencies and object-level semantic guidance, respectively. We concatenate them and use a lightweight mapper $G_{\omega}$ to predict a normalized soft affected-region map:
\begin{equation}
\hat{\mathbf{F}}_{\mathrm{eff}}
=
G_{\omega}
\left(
\left[
\mathbf{A}^{\mathrm{self}}_{\mathrm{obj}} ;
\mathbf{A}^{\mathrm{cross}}_{\mathrm{obj}}
\right]
\right).
\end{equation}
where $[\cdot; \cdot]$ denotes channel-wise concatenation. We implement $G_{\omega}$ with two per-pixel linear projections, a GELU activation, and a spatial softmax.

To supervise the predicted affected region, we construct a soft difference prior, denoted as $\mathbf{F}_{\mathrm{diff}}$, from the normalized distribution of the downsampled difference between the paired original video $\mathbf{V}_{\mathrm{ori}}$ and object-removed video $\mathbf{V}_{\mathrm{gt}}$. Finally, we define the Attention-guided Region Localization (ARL) loss as:
\begin{equation}
\mathcal{L}_{\mathrm{ARL}}
=
\mathrm{KL}
\left(
\mathbf{F}_{\mathrm{diff}}
\;\|\;
\hat{\mathbf{F}}_{\mathrm{eff}}
\right).
\end{equation}
This objective encourages the model's attention responses to cover both the target object and its affected regions, providing spatial guidance for more complete removal.

\subsection{Training Objective}
The overall objective combines the standard flow-matching loss with the proposed OIRD and ARL losses:
\begin{equation}
\mathcal{L}
=
\mathbb{E}_{t, \mathbf{z}_0, \boldsymbol{\epsilon}}
\left[
\left\lVert
\mathbf{v}
-
\hat{\mathbf{v}}
\right\rVert_2^2
+
\lambda_{\mathrm{OIRD}}
\mathcal{L}_{\mathrm{OIRD}}
+
\lambda_{\mathrm{ARL}}
\mathcal{L}_{\mathrm{ARL}}
\right],
\label{eq:training_loss}
\end{equation}

where $\lambda_{\mathrm{OIRD}}$ and $\lambda_{\mathrm{ARL}}$ are balancing weights, both set to $0.1$. Here, $\mathbf{v} = \mathrm{d}\mathbf{z}_t / \mathrm{d}t = \boldsymbol{\epsilon} - \mathbf{z}_0$ denotes the target velocity under the flow-matching parameterization, and $\hat{\mathbf{v}}$ denotes the velocity predicted by DiT.
\section{Experiments}

\begin{table*}[t]
\centering
\small
\setlength{\tabcolsep}{1.5pt}

\begin{tabular}{@{}l|ccc|ccc|ccc|ccc|c@{}}
\hline
\multirow{2}{*}{Method}
& \multicolumn{3}{c|}{ROSE-Bench}
& \multicolumn{3}{c|}{CAMERA-Bench}
& \multicolumn{3}{c|}{VOR-Eval}
& \multicolumn{3}{c|}{VOR-Wild}
& \multirow{2}{*}{Time (s)}
\\
\cline{2-13}

& PSNR $\uparrow$
& SSIM $\uparrow$
& LPIPS $\downarrow$
& PSNR $\uparrow$
& SSIM $\uparrow$
& LPIPS $\downarrow$
& PSNR $\uparrow$
& SSIM $\uparrow$
& LPIPS $\downarrow$
& Rem. $\uparrow$
& Mot. $\uparrow$
& User $\uparrow$
\\
\hline

FuseFormer~\cite{liu2021fuseformer}
& 25.9234 & 0.8857 & 0.2185
& 25.0572 & 0.9155 & 0.2020
& 21.0045 & 0.7641 & 0.3659
& 3.1860 & 3.1833 & 2.12 $\pm$ 0.25
& 9.998
\\

FGT~\cite{zhang2022flow}
& 26.2947 & 0.8999 & 0.1888
& 25.1073 & 0.9095 & 0.1905
& 21.0661 & 0.7734 & 0.3309
& 2.8942 & 2.8942 & 1.86 $\pm$ 0.18
& 96.241
\\

ProPainter~\cite{zhou2023propainter}
& 25.3245 & 0.9105 & 0.1281
& 25.4201 & 0.9203 & 0.1268
& 20.8422 & 0.7767 & 0.2630
& 2.7855 & 2.7855 & 1.84 $\pm$ 0.33
& 180.231
\\

VACE~\cite{jiang2025vace}
& 20.8162 & 0.7181 & 0.2238
& 16.9868 & 0.6928 & 0.2182
& 17.2086 & 0.5746 & 0.3488
& 3.7949 & 3.7897 & 1.18 $\pm$ 0.12
& 1191.391
\\

DiffuEraser~\cite{li2025diffueraser}
& 25.0410 & 0.8983 & 0.1198
& 26.7892 & 0.9409 & 0.0899
& 21.0845 & 0.7842 & 0.2503
& 1.3808 & 1.3808 & 1.35 $\pm$ 0.28
& 499.873
\\

MiniMax-Remover~\cite{zi2025minimax}
& 26.6838 & 0.9037 & 0.1490
& 26.6345 & 0.9324 & 0.0854
& 21.0583 & 0.7770 & 0.2762
& 3.9960 & 3.9960 & 2.98 $\pm$ 0.41
& 107.730
\\

VideoPainter~\cite{bian2025videopainter}
& 21.1890 & 0.8700 & 0.1823
& 15.2151 & 0.8378 & 0.1840
& 16.8494 & 0.7073 & 0.3359
& 2.9631 & 2.9574 & 1.18 $\pm$ 0.16
& 1733.691
\\

Gen-Omnimatte~\cite{lee2025generative}
& 26.6128 & 0.8977 & 0.1765
& 26.8216 & 0.9315 & 0.0853
& 21.5859 & 0.7734 & 0.2996
& 3.9728 & 3.9755 & 2.24 $\pm$ 0.31
& 245.783
\\

OmnimatteZero~\cite{samuel2025omnimattezero}
& 25.4721 & 0.8743 & 0.1835
& 23.8707 & 0.8965 & 0.1275
& 20.3396 & 0.7291 & 0.3049
& 3.7190 & 3.7135 & 2.41 $\pm$ 0.46
& 152.887
\\

ROSE~\cite{miao2025rose}
& 31.1299 & 0.9292 & 0.1193
& 26.4577 & 0.9286 & 0.0860
& 22.1554 & 0.7787 & 0.2605
& 4.0401 & 4.0374 & 2.71 $\pm$ 0.38
& 566.706
\\

YOSE~\cite{wu2026yose}
& 26.6063 & 0.9017 & 0.1554
& 26.8705 & 0.9371 & 0.0825
& 21.0740 & 0.7791 & 0.2831
& 3.8529 & 3.8556 & 2.70 $\pm$ 0.45
& 68.314
\\

EffectErase~\cite{fu2026effecterase}
& 26.9417 & 0.9048 & 0.1441
& 27.2018 & 0.9292 & 0.0849
& 22.3415 & 0.7737 & 0.2508
& 4.0491 & 4.0465 & 2.86 $\pm$ 0.31
& 431.872
\\

SVOR~\cite{hu2026ideal}
& 31.8744 & 0.9381 & 0.0985
& 27.2416 & 0.9435 & 0.0754
& 22.0942 & 0.7884 & 0.2294
& 4.1323 & 4.1349 & 3.21 $\pm$ 0.24
& 1873.936
\\
\hline

Ours
& \textbf{32.2982}
& \textbf{0.9478}
& \textbf{0.0978}
& \textbf{28.1384}
& \textbf{0.9462}
& \textbf{0.0718}
& \textbf{23.8686}
& \textbf{0.7944}
& \textbf{0.2104}
& \textbf{4.3288}
& \textbf{4.3315}
& \textbf{3.97 $\pm$ 0.22}
& 257.365
\\
\hline
\end{tabular}
\caption{
Quantitative comparison with state-of-the-art methods.
For VOR-Wild, Rem. and Mot. denote the Gemini-3 scores for
object-removal quality and temporal coherence, respectively, while User
denotes the mean user-study score for overall removal quality with 95\%
confidence intervals based on Student's $t$-distribution.
Additional results evaluated exclusively on the corresponding effect regions are provided in the supplementary material.
}
\label{tab:compare}
\end{table*}

\subsection{Implementation Details}
Our model is trained on ROSE~\cite{miao2025rose} and VOR~\cite{fu2026effecterase} datasets for 100 epochs with batch size 8, learning rate $1\times10^{-4}$, LoRA rank 64 and alpha 32. During training, video pairs are resized to $720\times480$ and sampled to 81 frames. Please refer to the supplementary material for more details.

\begin{figure*}[t]
    \centering
    \includegraphics[width=0.99\linewidth]{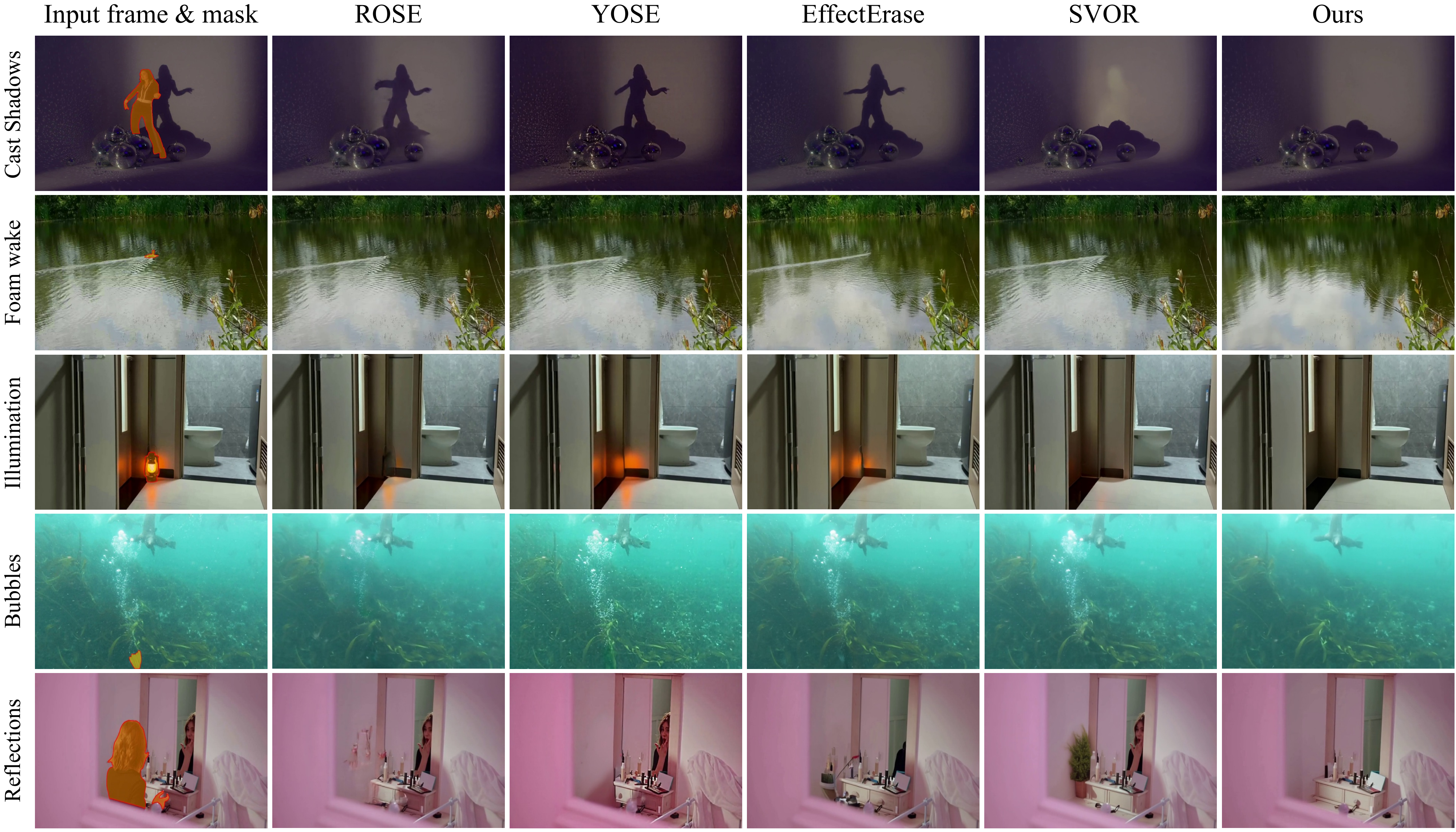}
    \caption{
    Qualitative comparison on diverse object-induced effects.
    Please refer to the supplementary material for more results.}
    \label{fig:main_comp}
\end{figure*}

\subsection{Evaluation Protocols}

\noindent\textbf{Baselines.}
We compare our method with the following approaches, which represent
the previous state-of-the-art in video inpainting and video object removal: FuseFormer~\cite{liu2021fuseformer}, FGT~\cite{zhang2022flow}, ProPainter~\cite{zhou2023propainter}, VACE~\cite{jiang2025vace}, DiffuEraser~\cite{li2025diffueraser}, MiniMax-Remover~\cite{zi2025minimax} , VideoPainter~\cite{bian2025videopainter}
, Gen-Omnimatte~\cite{lee2025generative}, OmnimatteZero~\cite{samuel2025omnimattezero}, ROSE~\cite{miao2025rose}, YOSE~\cite{wu2026yose}, EffectErase~\cite{fu2026effecterase} and SVOR~\cite{hu2026ideal}.

\vspace{1mm}
\noindent\textbf{Evaluation data.}
We use the following datasets:
\begin{enumerate}
    \item \textbf{ROSE-Bench}~\cite{miao2025rose} contains 60 video triplets synthesized with Unreal Engine. Each triplet includes an object-present video, the corresponding object-removed video, and the target object mask.

    \item \textbf{CAMERA-Bench.} We collect 40 realistic paired videos by capturing scenes with and without a controlled moving object. Each sample contains an input video, a SAM2-generated object mask~\cite{ravi2024sam}, and the object-removed ground truth. Although modest in scale, CAMERA-Bench complements existing benchmarks with real-world side-effect cases. Details are provided in the supplementary materials.

    \item \textbf{VOR-Eval}~\cite{fu2026effecterase} contains 43 paired video triplets. Each triplet includes an object-present video, the corresponding object-removed video, and the target object mask. The videos are sourced from both camera-captured scenes and 3D-rendered synthetic scenes.

    \item \textbf{VOR-Wild}~\cite{fu2026effecterase} is a test set consisting of 195 diverse real-world videos collected from the Internet. Each sample provides an input video and the corresponding target object mask.

\end{enumerate}

\noindent\textbf{Metrics.} For paired benchmarks (ROSE-Bench, CAMERA-Bench, VOR-Eval), we compute PSNR~\cite{hore2010image}, SSIM~\cite{wang2004image}, and LPIPS~\cite{zhang2018unreasonable} against the ground-truth videos. 

We further evaluate temporal consistency using flow-warping error~{\cite{lai2018learning}} computed by aligning consecutive predictions with optical flow estimated from the corresponding ground-truth frames; lower is better. Full details are provided in the supplementary material.

To more directly assess the removal of object-induced effects beyond the input mask, we additionally evaluate PSNR, SSIM, and LPIPS exclusively over the corresponding effect regions; the detailed protocol and results are provided in the supplementary material.

For the unpaired VOR-Wild benchmark, where ground-truth videos are unavailable, we use Gemini-3 to evaluate object removal quality and temporal coherence, reported as Removal and Motion scores, respectively. We further conduct a user study with 50 participants to rate the overall removal effect of each method. Details about the evaluation protocol are provided in the supplementary material.

\subsection{Results and Comparisons}
\noindent\textbf{Quantitative Comparison.}
As shown in Table~\ref{tab:compare}, our method achieves the best
PSNR, SSIM, and LPIPS on all paired benchmarks, indicating more accurate
pixel-level reconstruction, better preservation of background structures,
and improved perceptual similarity to the ground-truth videos. The
consistent gains across multiple benchmarks further suggest that our
method remains effective under diverse object categories, scene layouts,
and types of object-induced effects. On the unpaired VOR-Wild benchmark,
our method also obtains the highest Gemini-3 removal and motion scores,
together with the highest user-study rating. These results demonstrate
that the proposed method can more thoroughly eliminate target objects and
their associated effects, while preserving natural motion and producing
visually convincing reconstructions in challenging real-world videos.
The agreement between the automatic metrics and human evaluation further
supports the robustness and practical effectiveness of our approach.

We further evaluate temporal consistency using flow-warping
error~\cite{lai2018learning} on the paired benchmarks in
Table~\ref{tab:warp_error_main}. Our method consistently achieves the
lowest error across all three datasets, demonstrating that the
reconstructed regions remain more stable across adjacent frames. In
particular, the lower warping error indicates reduced temporal
flickering and fewer inconsistencies in background appearance and
structure. Combined with the superior reconstruction quality reported
in Table~\ref{tab:compare}, these results show that our method not only
removes target objects and their induced effects more completely, but
also produces visually plausible and temporally coherent background
reconstructions.

\begin{table}[t]
\centering
\footnotesize
\setlength{\tabcolsep}{4pt}

\footnotesize
{
\setlength{\tabcolsep}{3.5pt}
\begin{tabular}{@{}l|ccc@{}}
\hline
Method
& ROSE-Bench
& CAMERA-Bench
& VOR-Eval \\
\hline

FuseFormer~\cite{liu2021fuseformer}
& 0.004234 & 0.002088 & 0.006063 \\

FGT~\cite{zhang2022flow}
& 0.004912 & 0.001831 & 0.007903 \\

ProPainter~\cite{zhou2023propainter}
& 0.004444 & 0.003563 & 0.007552 \\

VACE~\cite{jiang2025vace}
& 0.004515 & 0.006747 & 0.009817 \\

DiffuEraser~\cite{li2025diffueraser}
& 0.004415 & 0.002339 & 0.007184 \\

MiniMax-Remover~\cite{zi2025minimax} 
& 0.004032 & 0.002432 & 0.006304 \\

VideoPainter~\cite{bian2025videopainter}
& 0.004930 & 0.005793 & 0.009890 \\

Gen-Omnimatte~\cite{lee2025generative}
& 0.003847 & 0.000950 & 0.005870 \\

OmnimatteZero~\cite{samuel2025omnimattezero}
& 0.003792 & 0.000940 & 0.006261 \\

ROSE~\cite{miao2025rose}
& 0.004166 & 0.002350 & 0.006877 \\

YOSE~\cite{wu2026yose}
& 0.003955 & 0.001316 & 0.006056 \\

EffectErase~\cite{fu2026effecterase}
& 0.003985 & 0.001728 & 0.006012 \\

SVOR~\cite{hu2026ideal}
& 0.003993 & 0.001325 & 0.007463 \\
\hline

Ours
& \textbf{0.003734}
& \textbf{0.000926}
& \textbf{0.005122} \\
\hline
\end{tabular}
}
\caption{
Temporal consistency comparison with state-of-the-art methods using flow-warping error. Lower is better.
}
\label{tab:warp_error_main}
\end{table}

\vspace{1mm}
\noindent\textbf{Qualitative Comparison.} As shown in Figure~\ref{fig:main_comp}, our method consistently achieves more complete removal of both target objects and their induced effects across diverse scenarios. Existing methods can often erase the main body of the target object, but still leave visible remnants or fail to suppress effects extending beyond the input mask, such as shadows, reflections, illumination changes, and water-surface distortions. In contrast, our method more effectively removes both the target object and its associated effects, while reconstructing backgrounds that are more consistent with the surrounding spatial context. More comparisons with additional baselines are provided in the supplementary material.

\subsection{Ablation Studies}
We conduct ablation studies to evaluate the effectiveness of Object-Induced Relation Distillation (OIRD), Object-aware Framewise Context Cross-Attention (OFCCA), and Attention-guided Region Localization (ARL) on VOR-Eval. As shown in Table~\ref{tab:ablation}, each component improves the final performance, and combining all three components achieves the best overall results. 

\begin{figure}[t]
\centering
\includegraphics[width=0.99\columnwidth]{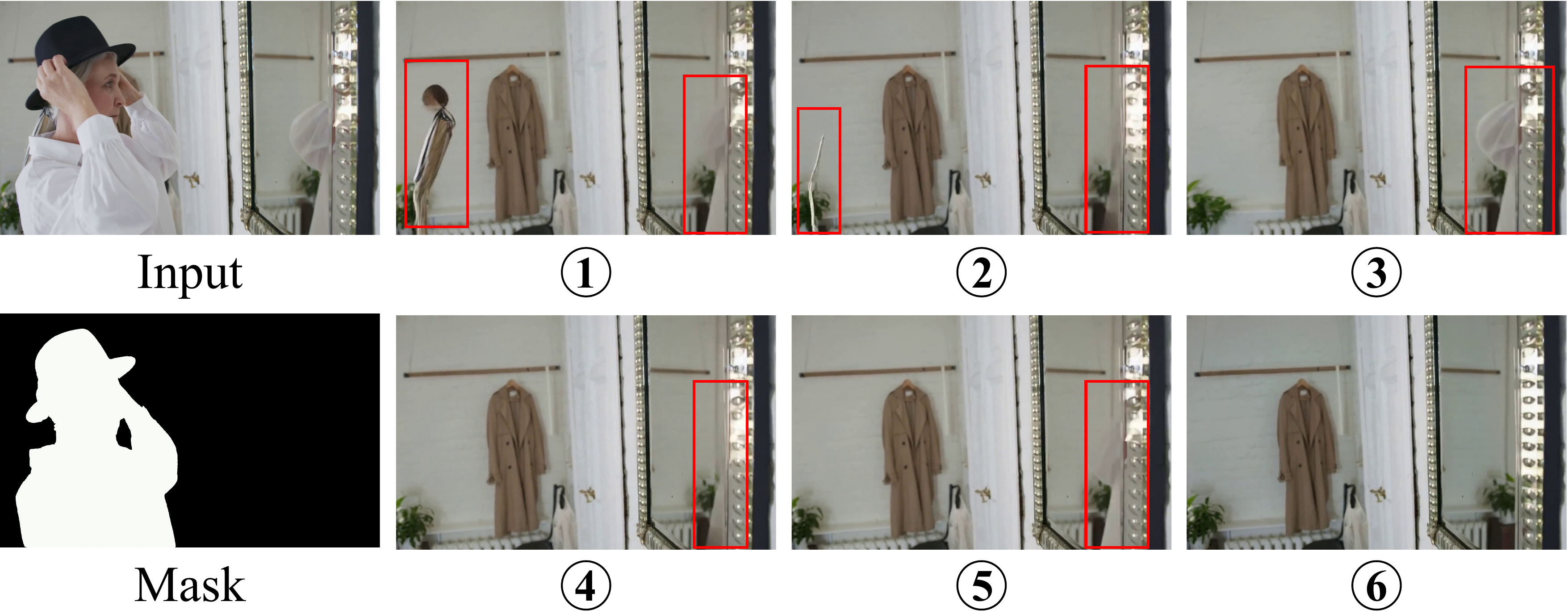}
\caption{
Qualitative results of different settings.
}
\label{fig:aba_com}
\end{figure}

Figure~\ref{fig:aba_com} further provides qualitative comparisons corresponding to the settings in Table~\ref{tab:ablation}. The baseline in~{\large\ding{172}} leaves obvious residual hair and mirror reflections. Using only OIRD in~{\large\ding{173}} improves awareness of target-related effects and suppresses part of the reflection, but human remnants and reflected artifacts still remain. Adding OFCCA in setting~{\large\ding{174}} improves the completeness
of human-body removal, although the reflected target remains noticeable. The variants in~{\large\ding{175}}  and~{\large\ding{176}} further improve the removal of the main body, yet they still fail to completely suppress the reflection in the mirror. In contrast, the full model in~{\large\ding{177}} combines OIRD, OFCCA, and ARL, achieving the most complete removal of both the target person and the mirror-induced residual effects.

\begin{table}[t]
\centering
\small
\setlength{\tabcolsep}{3.5pt}

\begin{tabular}{ccc|ccc|c}
\hline
OIRD & OFCCA & ARL
& PSNR $\uparrow$
& SSIM $\uparrow$
& LPIPS $\downarrow$
& No. \\
\hline

           &            &            & 22.3897 & 0.7866 & 0.2439 & \ding{172} \\
DINOv2 &            &            & 22.7473 & 0.7873 & 0.2337 & \ding{173} \\
           & \checkmark &            & 22.4818 & 0.7869 & 0.2363 & \ding{174} \\
           & \checkmark & \checkmark & 22.8686 & 0.7890 & 0.2256 & \ding{175} \\
DINOv2 & \checkmark &            & 22.9477 & 0.7904 & 0.2235 & \ding{176} \\
\hline
 SAM2 & \checkmark & \checkmark & 23.3398 & 0.7911 & 0.2208 & \\ 
 V-JEPA2 & \checkmark & \checkmark & 23.0683 & 0.7892 & 0.2223 & \\
\hline
DINOv2 & \checkmark & \checkmark
& \textbf{23.8686}
& \textbf{0.7944}
& \textbf{0.2104}
& \ding{177} \\
\hline
\end{tabular}

\caption{
Ablation study of different settings.
}
\label{tab:ablation}
\end{table}

\begin{figure}[t]
\centering
\includegraphics[width=0.99\columnwidth]{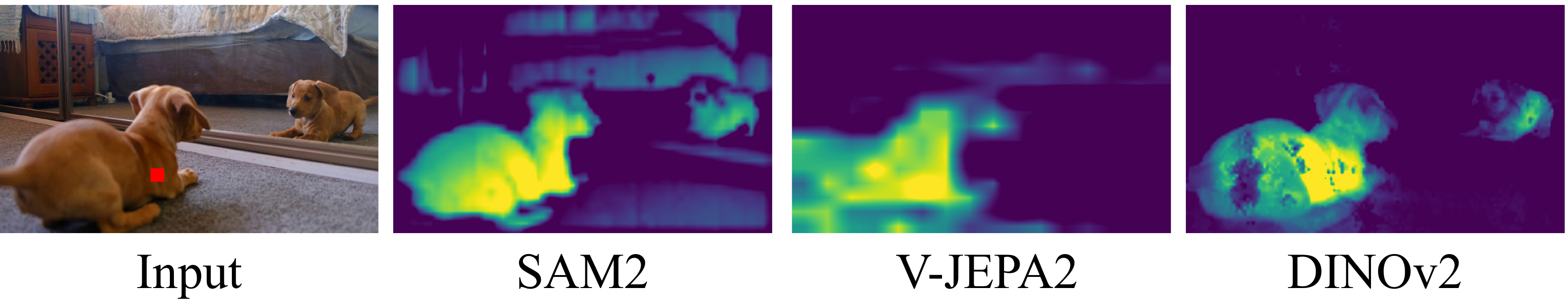}
\caption{
Visualization of attention weights from all query tokens to selected key tokens in different VFMs. Compared with SAM2 and V-JEPA2, DINOv2 yields more localized and semantically consistent responses over the target object and its associated effect regions.
}
\label{fig:tea_com}
\end{figure}

\begin{figure}[t]
\centering
\includegraphics[width=0.99\columnwidth]{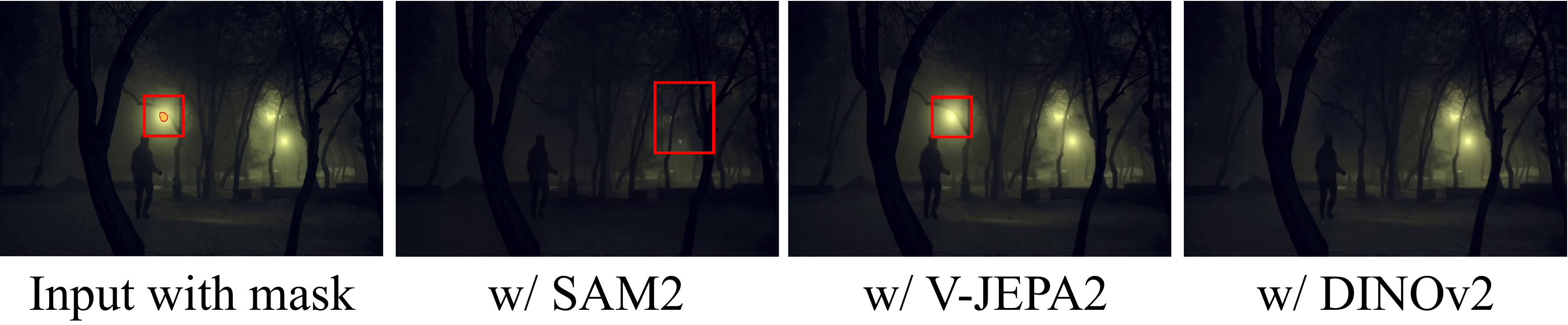}
\caption{
Qualitative comparison using different teacher VFMs in OIRD. DINOv2 yields more accurate and localized removal, whereas
SAM2 may mistakenly remove other objects of the same category and
V-JEPA2 often fails to completely eliminate the target object and its
induced effects.
}
\label{fig:tea_com_res}
\end{figure}

We also study the effect of using different VFMs as teachers in OIRD, including SAM2~\cite{ravi2025sam}, V-JEPA2~\cite{assran2025v}, and DINOv2.
We visualize their responses in Figure~\ref{fig:tea_com}. SAM2 often produces high responses on regions weakly related to the target object, while V-JEPA2 lacks precise token-level localization of the object and its side effects.
In contrast, DINOv2 provides more localized and semantically consistent object--effect responses.
Correspondingly, in Figure~\ref{fig:tea_com_res} and Table~\ref{tab:ablation}, DINOv2 leads to more complete side-effect removal and better scores than SAM2 and V-JEPA2.
However, SAM2 and V-JEPA2 still outperform setting {\large\ding{175}} in Table~\ref{tab:ablation}, showing the effectiveness of introducing VFM in OIRD.

\section{Conclusion}
 In this paper, we present an understanding-centric video object removal framework that enhances diffusion-based erasing with external object--effect knowledge and internal contextual and spatial guidance. Specifically, we introduce Object-Induced Relation Distillation to transfer fine-grained object--effect relations, Object-aware Framewise Context Cross-Attention to integrate target semantics with frame-specific background context, and Attention-guided Region Localization to improve spatial awareness of target-related affected regions. By combining these components, our method more completely removes target objects and their induced effects while preserving background fidelity and spatio-temporal consistency. Extensive experiments demonstrate leading performance across multiple benchmarks.

\clearpage
{
    \small
    \bibliographystyle{ieeenat_fullname}
    \bibliography{main}
}

\clearpage
\appendix
\section{CAMERA-Bench}
\begin{figure}[t]
\centering
\includegraphics[width=0.9\columnwidth]{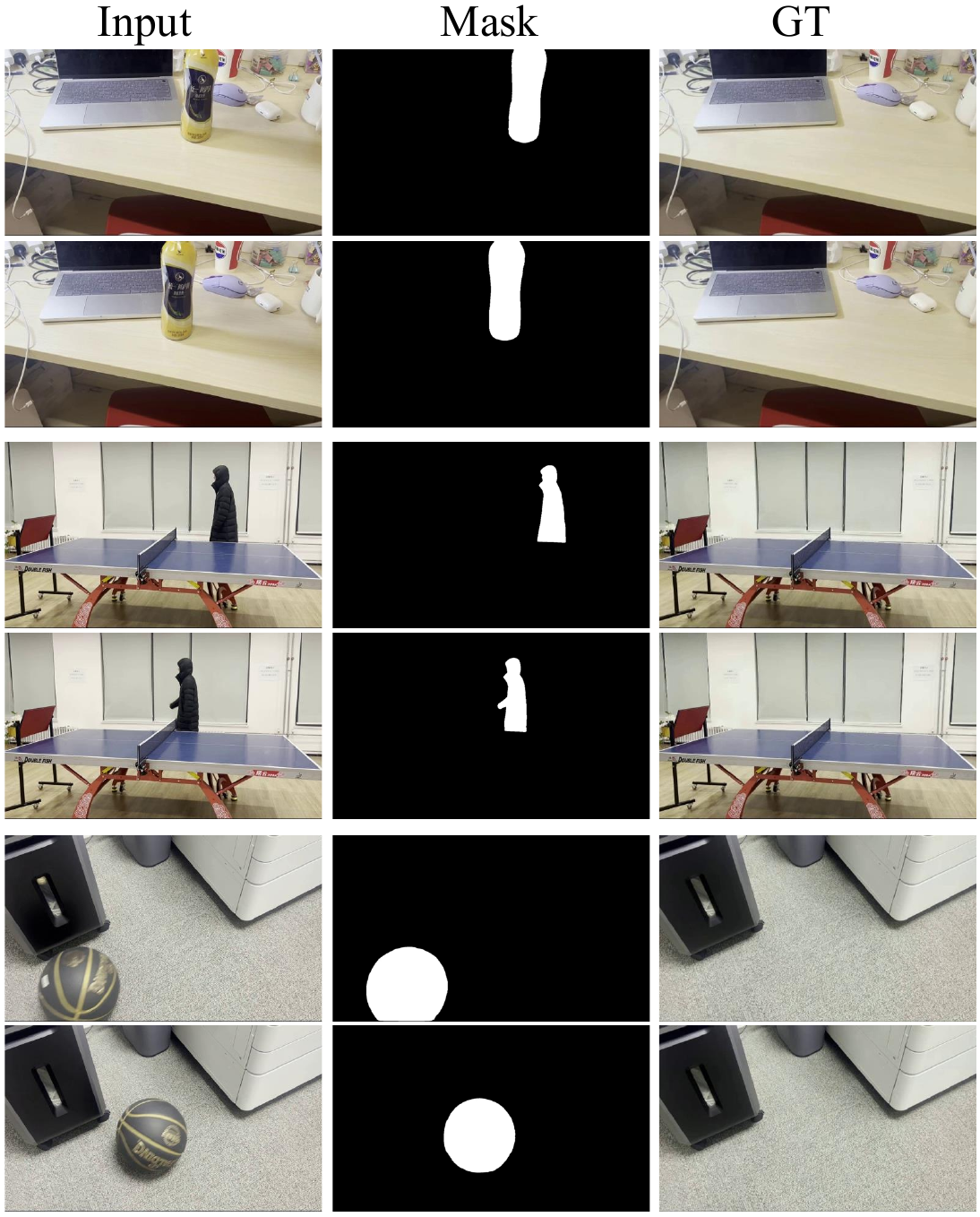}
\caption{
Samples from CAMERA-Bench for video object removal.
}
\label{fig:camera_bench}
\end{figure}

We construct CAMERA-Bench by capturing paired real-world videos of the same scene with and without a controlled moving object. For each pair, the object-present and object-absent videos are temporally trimmed to 81 frames. All videos are captured under consistent settings and have a spatial resolution of ($1920 \times 1080$). Object masks are generated using SAM2 and manually inspected and corrected when necessary. All individuals appearing in CAMERA-Bench provided informed consent for data collection and research use. CAMERA-Bench contains 40 paired videos covering representative object-induced effects and mixed interactions. The benchmark spans diverse scenes, moving-object categories, and background conditions. Examples are shown in Figure ~\ref{fig:camera_bench}.

\section{Additional Evaluation Details}

\noindent\textbf{Effect-Region Evaluation.}
To directly evaluate the removal of object-induced effects beyond the input object mask, we additionally report effect-region PSNR and SSIM, together with an effect-focused LPIPS score. Specifically, we construct a binary difference mask by thresholding the per-pixel $\ell_2$ distance between the object-present input video and its object-absent ground-truth counterpart. We then exclude the target-object mask from the difference mask to obtain the object-induced effect region outside the input mask. PSNR and SSIM are computed only over pixels within this region. For LPIPS, we replace pixels outside the effect region in each prediction with the corresponding ground-truth pixels before computing the perceptual distance, thereby eliminating pixel-level differences outside the evaluated region. All methods are evaluated using the same effect masks and evaluation protocol. As shown in Table~\ref{tab:effect_region}, our method achieves more accurate reconstruction of regions affected by shadows, reflections, illumination changes, and other object-induced effects beyond the input mask.

\begin{table*}[t]
\centering
\setlength{\tabcolsep}{3.5pt}
{
\begin{tabular}{l|ccc|ccc|ccc}
\hline
\multirow{2}{*}{Method}
& \multicolumn{3}{c|}{ROSE-Bench}
& \multicolumn{3}{c|}{CAMERA-Bench}
& \multicolumn{3}{c}{VOR-Eval} \\
\cline{2-10}
& PSNR $\uparrow$
& SSIM $\uparrow$
& LPIPS $\downarrow$
& PSNR $\uparrow$
& SSIM $\uparrow$
& LPIPS $\downarrow$
& PSNR $\uparrow$
& SSIM $\uparrow$
& LPIPS $\downarrow$ \\
\hline

FuseFormer~\cite{liu2021fuseformer}
& 17.1780 & 0.6637 & 0.0813
& 19.7633 & 0.7855 & 0.0773
& 17.6084 & 0.6329 & 0.2650 \\

FGT~\cite{zhang2022flow}
& 17.1951 & 0.6728 & 0.0805
& 19.6013 & 0.7947 & 0.0768
& 17.5656 & 0.6433 & 0.2603 \\

ProPainter~\cite{zhou2023propainter}
& 16.9533 & 0.6691 & 0.0777
& 19.7189 & 0.8212 & 0.0702
& 17.3333 & 0.6480 & 0.2393 \\

VACE~\cite{jiang2025vace}
& 15.8542 & 0.5914 & 0.0903
& 16.6771 & 0.6206 & 0.0980
& 16.3969 & 0.5219 & 0.2847 \\

DiffuEraser~\cite{li2025diffueraser}
& 17.0905 & 0.6582 & 0.0774
& 20.0279 & 0.8295 & 0.0692
& 17.3647 & 0.6479 & 0.2369 \\

MiniMax-Remover~\cite{zi2025minimax}
& 17.4385 & 0.6775 & 0.0798
& 20.2539 & 0.8235 & 0.0689
& 17.5684 & 0.6416 & 0.2552 \\

VideoPainter~\cite{bian2025videopainter}
& 15.7812 &0.5829 & 0.0849
& 18.3355 & 0.7649 & 0.0848
& 17.3737 & 0.6317 & 0.2602 \\

Gen-Omnimatte~\cite{lee2025generative}
& 19.5182 & 0.7229 & 0.0788
& 23.7920 & 0.8502 & 0.0667
& 18.5688 & 0.6598 & 0.2541 \\

OmnimatteZero~\cite{samuel2025omnimattezero}
& 18.4218 & 0.6834 & 0.0820
& 20.6756 & 0.8028 & 0.0765
& 17.9424 & 0.6266 & 0.2656 \\

ROSE~\cite{miao2025rose}
& 25.4045 & 0.8238 & 0.0601
& 22.4831 & 0.8352 & 0.0663
& 19.3843 & 0.6654 & 0.2467 \\

YOSE~\cite{wu2026yose}
& 17.4409 & 0.6747 & 0.0806
& 20.4666 & 0.8293 & 0.0705
& 17.4788 & 0.6440 & 0.2569 \\

EffectErase~\cite{fu2026effecterase}
& 20.9648 & 0.7529 & 0.0770
& 24.1832 & 0.8575 & 0.0650
& 20.6618 & 0.6736 & 0.2375 \\

SVOR~\cite{hu2026ideal}
& 26.9857 & 0.8524 & 0.0532
& 23.4310 & 0.8641 & 0.0674
& 19.3223 & 0.6760 & 0.2394 \\
\hline

Ours
& \textbf{27.0787} & \textbf{0.8552} & \textbf{0.0520}
& \textbf{25.0616} & \textbf{0.8709} & \textbf{0.0604}
& \textbf{22.2544} & \textbf{0.6984} & \textbf{0.2063} \\
\hline
\end{tabular}
}
\caption{
Quantitative evaluation on object-induced effect regions outside the input object mask on ROSE-Bench, CAMERA-Bench, and VOR-Eval.
PSNR and SSIM are computed exclusively over the effect-region pixels, while LPIPS denotes the effect-focused perceptual distance.
}
\label{tab:effect_region}
\end{table*}

\noindent\textbf{Temporal Consistency Evaluation.}
We evaluate temporal consistency on the three paired benchmarks using
flow-warping error. Let $\hat{\mathbf{I}}_t$ and
$\hat{\mathbf{I}}_{t+1}$ denote two consecutive predicted frames. We
estimate bidirectional optical flow between the corresponding
ground-truth object-absent frames using the pretrained RAFT-Large model~\cite{RAFT},
and denote the forward flow by
$\mathbf{F}_{t\rightarrow t+1}$. Using backward sampling, we warp
$\hat{\mathbf{I}}_{t+1}$ to the coordinate system of
$\hat{\mathbf{I}}_t$. The frame-pair warping error is defined as
\begin{equation}
E_t^{\mathrm{warp}}
=
\frac{
\sum_{\mathbf{p}}
\mathbf{M}_t^{\mathrm{valid}}(\mathbf{p})
\left\|
\hat{\mathbf{I}}_t(\mathbf{p})
-
\mathcal{W}\!\left(
\hat{\mathbf{I}}_{t+1},
\mathbf{F}_{t\rightarrow t+1}
\right)(\mathbf{p})
\right\|_1
}{
3\sum_{\mathbf{p}}
\mathbf{M}_t^{\mathrm{valid}}(\mathbf{p})
},
\label{eq:flow_warping}
\end{equation}
where $\mathcal{W}$ denotes backward warping with bilinear interpolation,
and $\mathbf{M}_t^{\mathrm{valid}}$ excludes out-of-boundary pixels and
unreliable flow regions identified through forward--backward
consistency. RGB values are normalized to $[0,1]$, and frame pairs with
fewer than 100 valid pixels are discarded. We first average the valid
frame-pair errors within each video and then report the macro average
over all valid videos in each benchmark. The same optical-flow estimator
and evaluation protocol are used for all methods.

\begin{table}[t]
\centering
\setlength{\tabcolsep}{3.8pt}
\begin{tabular}{cc|ccc}
\hline
$\lambda_{\mathrm{OIRD}}$
& $\lambda_{\mathrm{ARL}}$
& PSNR $\uparrow$
& SSIM $\uparrow$
& LPIPS $\downarrow$ \\
\hline

0
& 0.1
& 22.8686
& 0.7890
& 0.2256 \\

0.05
& 0.1
& 23.4215
& 0.7922
& 0.2178 \\

0.1
& 0.1
& \textbf{23.8686}
& \textbf{0.7944}
& \textbf{0.2104} \\

0.2
& 0.1
& 23.6127
& 0.7931
& 0.2149 \\
\hline

0.1
& 0
& 22.9477
& 0.7904
& 0.2235 \\

0.1
& 0.05
& 23.5068
& 0.7928
& 0.2162 \\

0.1
& 0.2
& 23.6412
& 0.7933
& 0.2138 \\
\hline
\end{tabular}
\caption{
Sensitivity analysis of the loss weights
$\lambda_{\mathrm{OIRD}}$ and $\lambda_{\mathrm{ARL}}$ on VOR-Eval.
When varying one weight, the other is fixed at 0.1.
}
\label{tab:loss_weight_sensitivity}
\end{table}

\noindent\textbf{Sensitivity to Loss Weights.}
We study the sensitivity of the balancing weights
$\lambda_{\mathrm{OIRD}}$ and $\lambda_{\mathrm{ARL}}$ by varying one
weight while fixing the other at 0.1. As shown in
Table~\ref{tab:loss_weight_sensitivity}, removing either auxiliary
objective results in a clear performance degradation. Increasing each
weight from 0 to a moderate value improves reconstruction quality,
whereas further increasing it to 0.2 provides no additional benefit
and slightly degrades performance. This suggests that excessively
strong auxiliary supervision may over-constrain the primary denoising
objective. The model achieves the best overall PSNR, SSIM, and LPIPS
when both weights are set to 0.1. We therefore use
$\lambda_{\mathrm{OIRD}}=\lambda_{\mathrm{ARL}}=0.1$
in all experiments.

\section{More Implementation Details}
\noindent\textbf{Computing Environment.} All experiments are conducted on a server equipped with eight NVIDIA H20 GPUs, each with approximately 96 GB of GPU memory, two AMD EPYC 9K84 96-Core processors, and 2.2 TiB of system memory. The server runs Ubuntu 22.04.4 LTS with NVIDIA driver 575.57.08. Our implementation uses PyTorch 2.9.0, CUDA 12.8, and cuDNN 9.10.2. The main software dependencies include Diffusers 0.36.0, Transformers 4.57.3, and Accelerate 1.11.0.

\noindent\textbf{Baseline and Inference Settings.} For all baseline methods, we use the officially released implementations and pretrained checkpoints, following the recommended inference settings provided by the corresponding authors. For our method, we use a fixed random seed of 43 for all inference experiments. Each method is run once for each input video under the same evaluation protocol.

\begin{figure}[t]
\centering
\includegraphics[width=0.9\columnwidth]{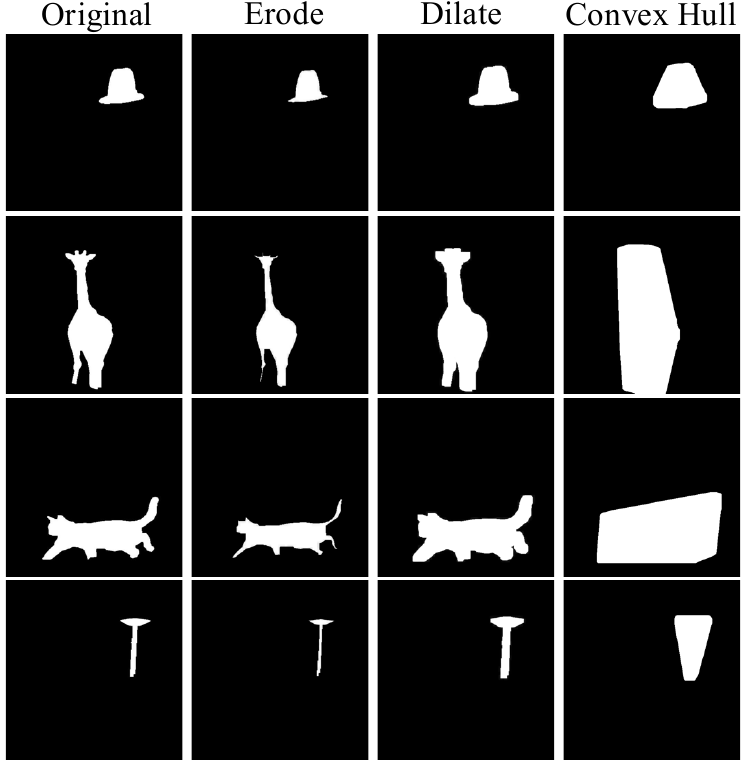}
\caption{Visualization of mask augmentation variants.}
\label{fig:mask_enhance}
\end{figure}

\noindent\textbf{Mask Augmentation.}
In real-world scenarios, user-provided masks are often coarse and lack precise alignment with object contours. Relying exclusively on precise masks during training leads to a notable discrepancy from the masks encountered at inference time. To improve robustness, we incorporate a mask augmentation scheme that generates four variants of the ground-truth object mask, enabling the model to better handle imperfect user inputs. Specifically, as shown in Figure ~\ref{fig:mask_enhance}, we use four mask types to augment the object mask:
(1) Original mask, which provides an accurate delineation of the object.
(2) Eroded mask, produced via morphological erosion to emulate cases where the user-provided mask under-covers the object.
(3) Dilated mask, obtained through morphological dilation to mimic over-segmented user inputs.
(4) Convex-hull mask, formed by computing the convex hull of the original mask, applying a slight morphological expansion, and performing a final smoothing.

\section{Limitations} 
\begin{figure}[t]
\centering
\includegraphics[width=0.9\columnwidth]{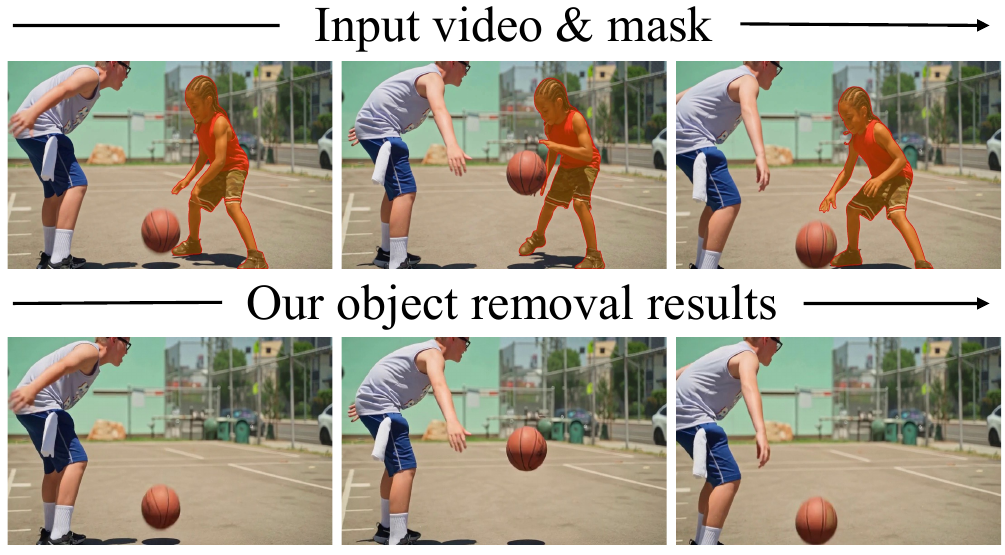}
\caption{
Limitation in scenes with strong physical interactions.
After removing the person, the basketball may still follow its original
trajectory, resulting in physically implausible motion.
}
\label{fig:limitation}
\end{figure}

Our method mainly focuses on removing target objects and their induced visual effects. However, it may still be limited in scenes with strong physical interactions. As shown in Figure ~\ref{fig:limitation}, after removing a person playing basketball, the ball may continue moving along its original trajectory, resulting in physically implausible motion. This suggests that modeling high-level object interaction and physical causality remains an important direction for future work.

\section{Evaluation Protocol on VOR-Wild}

\begin{figure*}[t]
    \centering
    \includegraphics[width=0.7\linewidth]{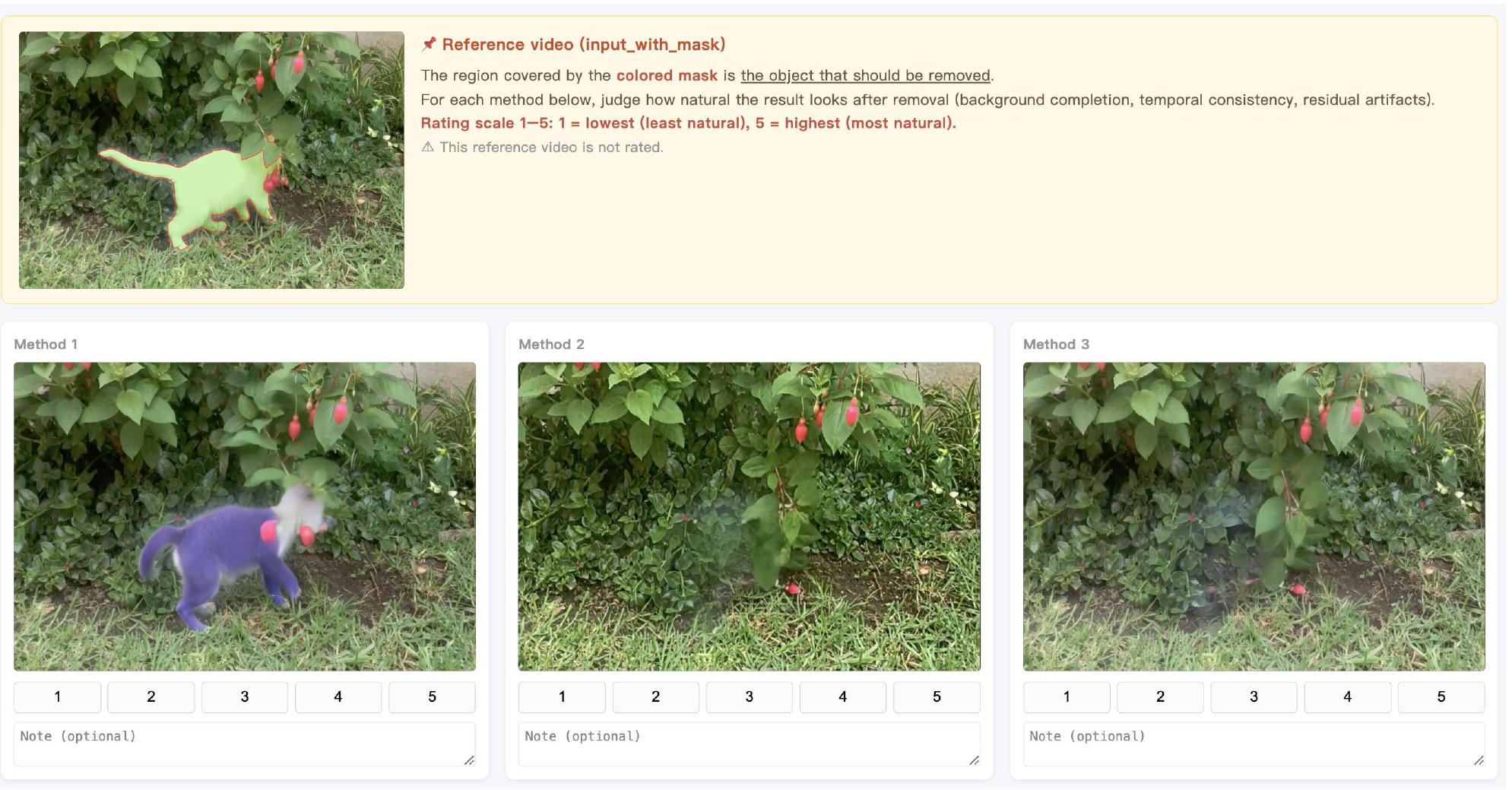}
    \caption{Illustration of the user-study interface. Due to space limitations,
    only three anonymized results are displayed, whereas the actual study includes all evaluated methods.}
    \label{fig:user}
\end{figure*}

\noindent\textbf{Human User Study.}
As shown in Figure ~\ref{fig:user}, we design a user study to assess the perceived quality of video object removal results. Each participant is shown the masked input video, where the colored mask indicates the target object to be removed, together with the object-removed results from different methods. Participants are asked to rate each result on a 1--5 scale, where higher scores indicate better removal quality. The rating considers whether the target object is completely removed, whether the background is plausibly completed, whether residual artifacts remain, and whether the edited video is temporally consistent. Method identities are hidden from the participants, and the presentation order of different methods is randomized for each sample. All participants receive the same rating instructions and are allowed to replay the videos before assigning scores. We report the average rating across all participants and evaluated videos as the final user-study score.

\begin{figure*}[t]
    \centering
    \includegraphics[width=0.7\linewidth]{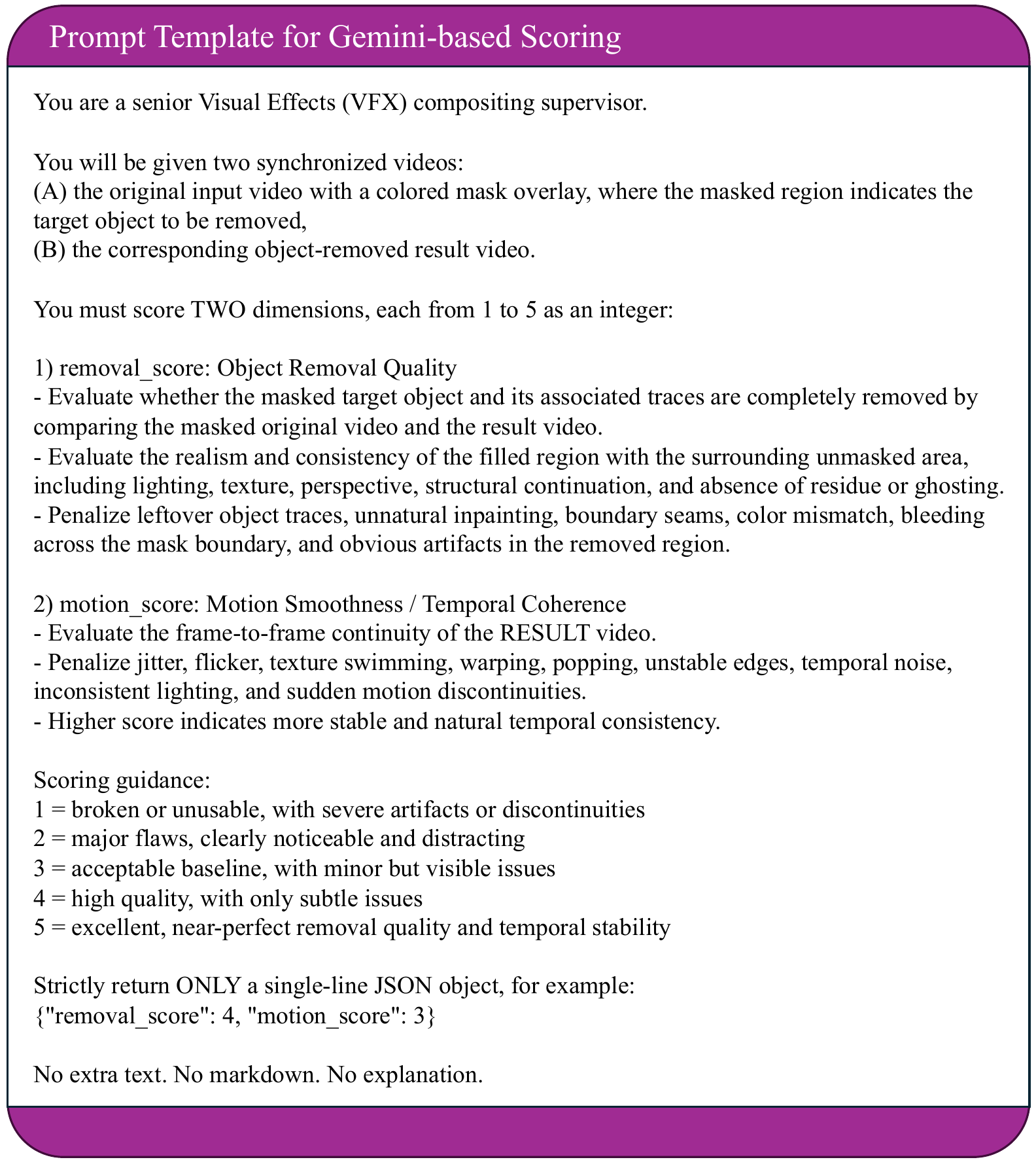}
    \caption{Prompt used for Gemini-3-based evaluation on VOR-Wild.}
    \label{fig:gemini}
\end{figure*}

\noindent\textbf{Gemini-based Automatic Evaluation.}
Since VOR-Wild does not provide ground-truth object-removed videos, we adopt Gemini-3 as an automatic evaluator to assess removal quality on this benchmark. As shown in Figure~\ref{fig:gemini}, Gemini takes the binary mask video and the corresponding object-removed result as inputs, and is asked to score two dimensions from 1 to 5. The Removal score evaluates mask-guided removal completeness and visual realism, including residue, boundary artifacts, and background consistency. The Motion score evaluates temporal smoothness and coherence, including flickering, jittering, and unstable textures. We use the same evaluation prompt and model configuration for all methods, without providing method identities to the evaluator. All results are evaluated under identical video resolution and duration settings. The final Removal and Motion scores are obtained by averaging the scores over all VOR-Wild samples.

\section{More Visualization of Comparison}
\begin{figure*}[t]
    \centering
    \includegraphics[width=0.99\linewidth]{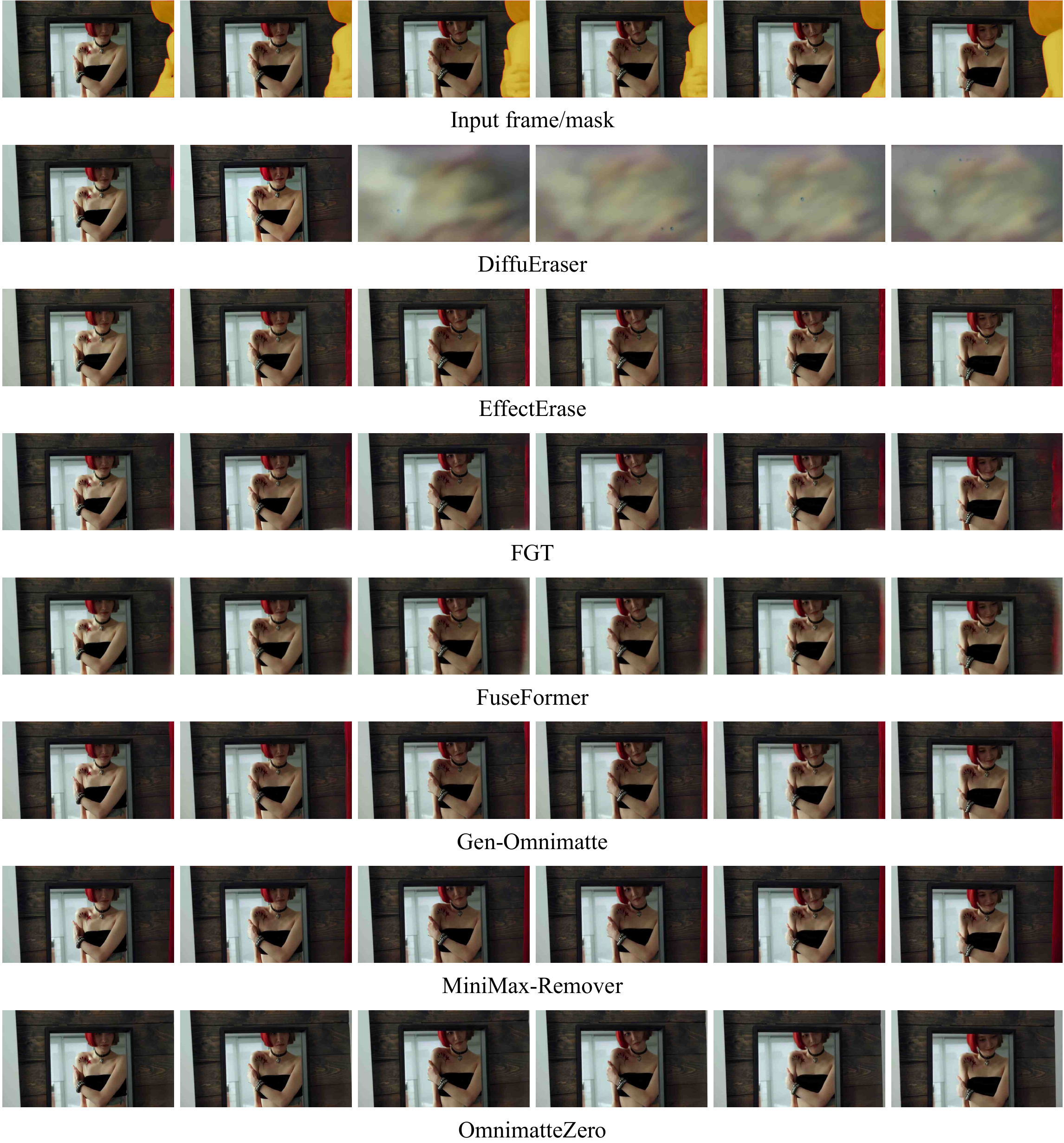}
    \caption{More comparison results on VOR-Wild.(1/2)}
    \label{fig:supp_comp_1}
\end{figure*}

\begin{figure*}[t]
    \centering
    \includegraphics[width=0.99\linewidth]{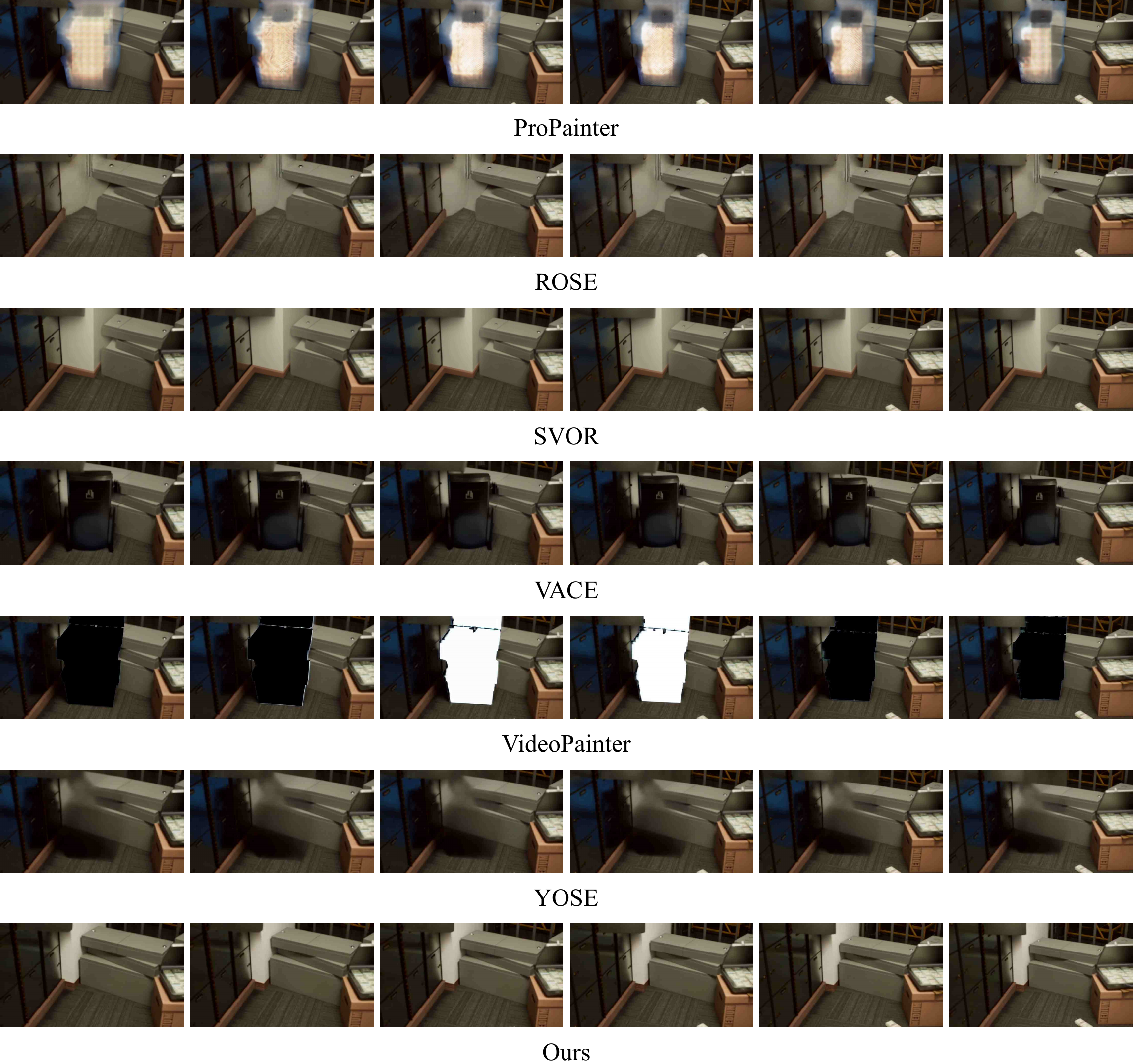}
    \caption{More comparison results on ROSE-Bench.(2/2)}
    \label{fig:supp_comp_14}
\end{figure*}

Figures~\ref{fig:supp_comp_1}--\ref{fig:supp_comp_14} present additional qualitative comparisons with all baseline methods across diverse real-world scenarios. These examples cover challenging object-induced effects, including shadows, reflections, water-surface distortions. Existing methods can generally remove the main body of the target object, but often leave visible object remnants or fail to eliminate effects extending beyond the input mask. Some methods also introduce noticeable boundary artifacts, inconsistent textures, or temporally unstable reconstruction in the removed regions. In contrast, our method more completely removes both the target object and its associated effects, while reconstructing visually plausible backgrounds with improved spatial and temporal consistency.

\begin{figure*}[t]
    \centering
    \includegraphics[width=0.99\linewidth]{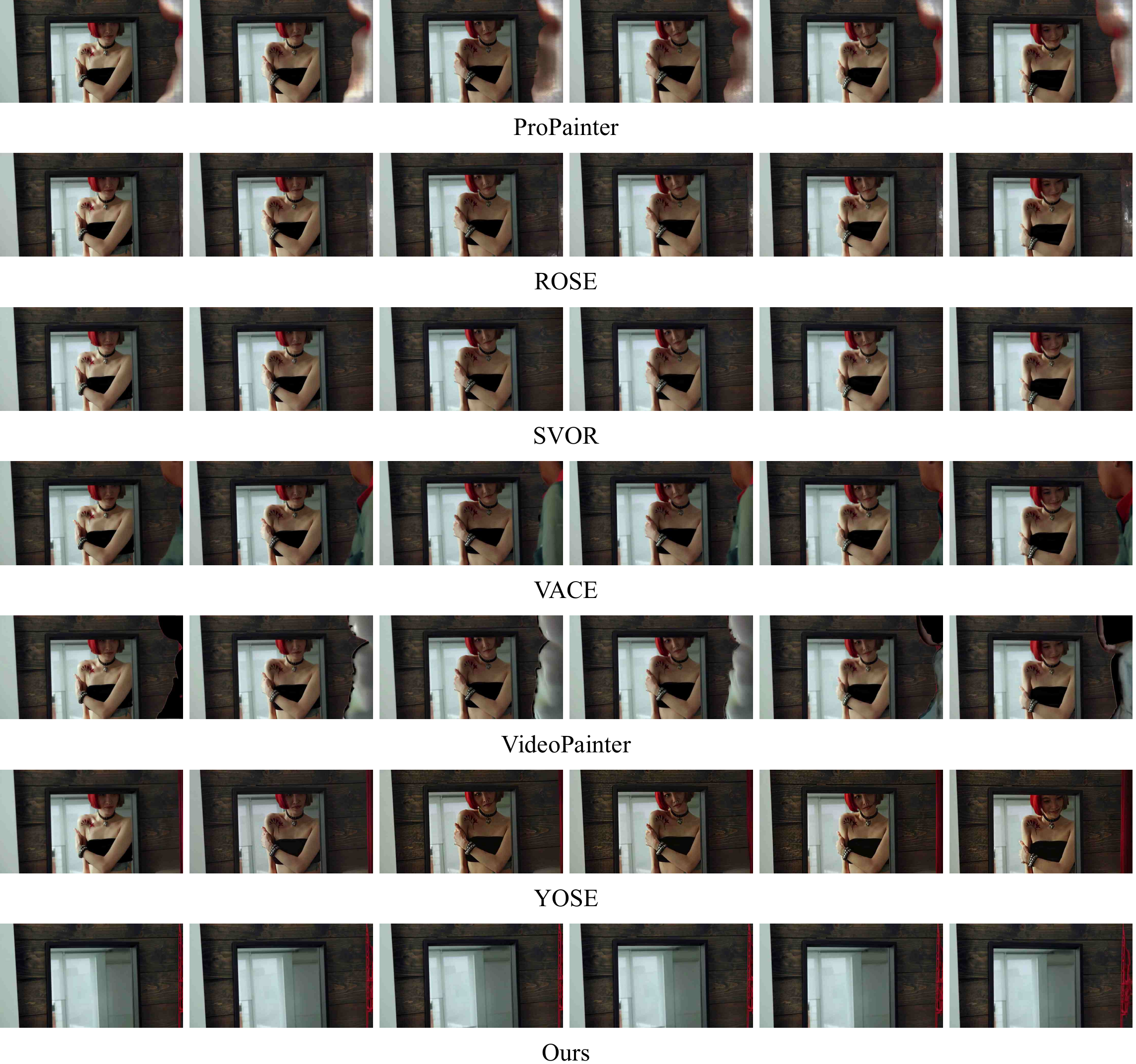}
    \caption{More comparison results on VOR-Wild.(2/2)}
\end{figure*}

\begin{figure*}[t]
    \centering
    \includegraphics[width=0.99\linewidth]{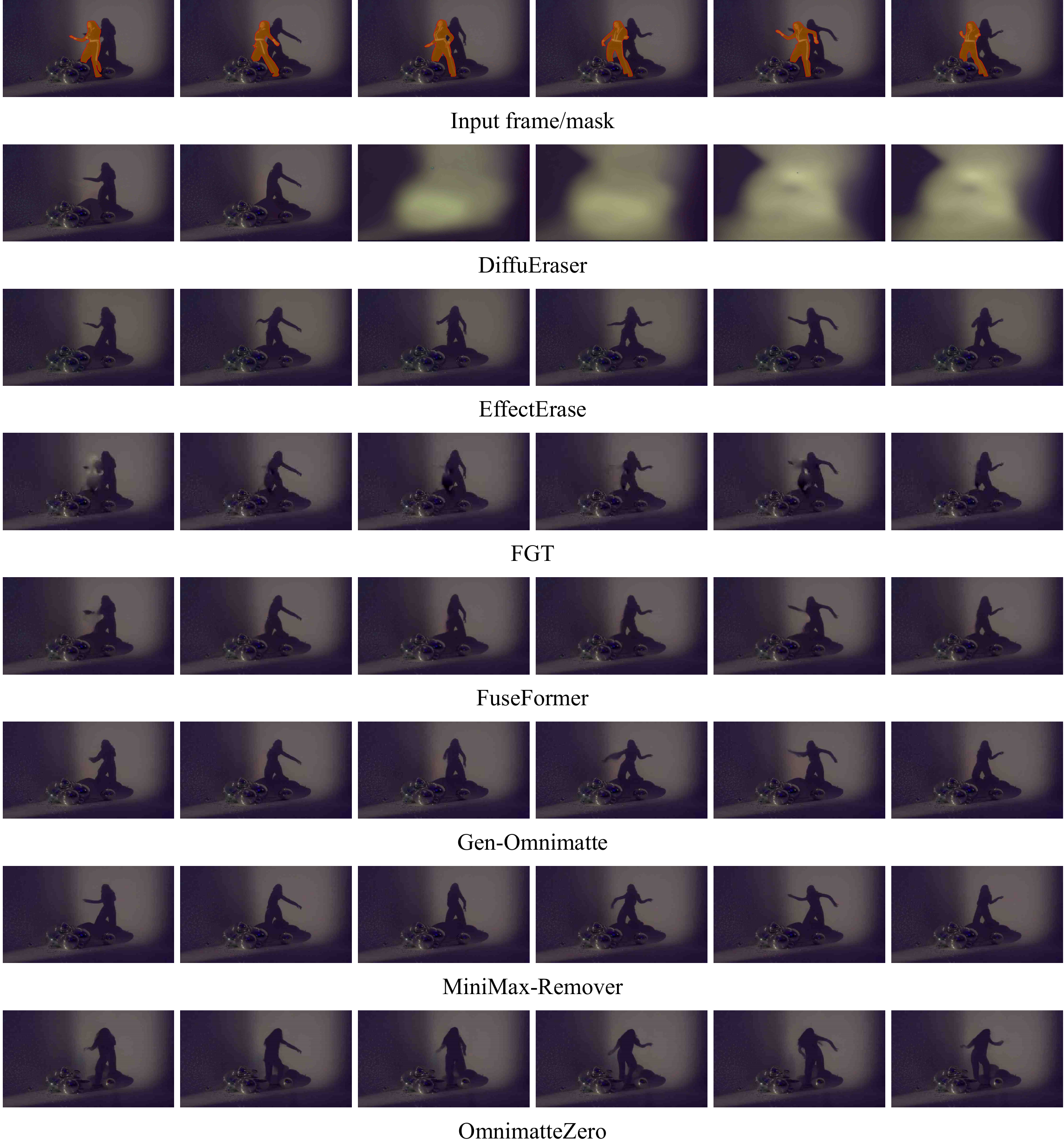}
    \caption{More comparison results on VOR-Wild.(1/2)}
\end{figure*}

\begin{figure*}[t]
    \centering
    \includegraphics[width=0.99\linewidth]{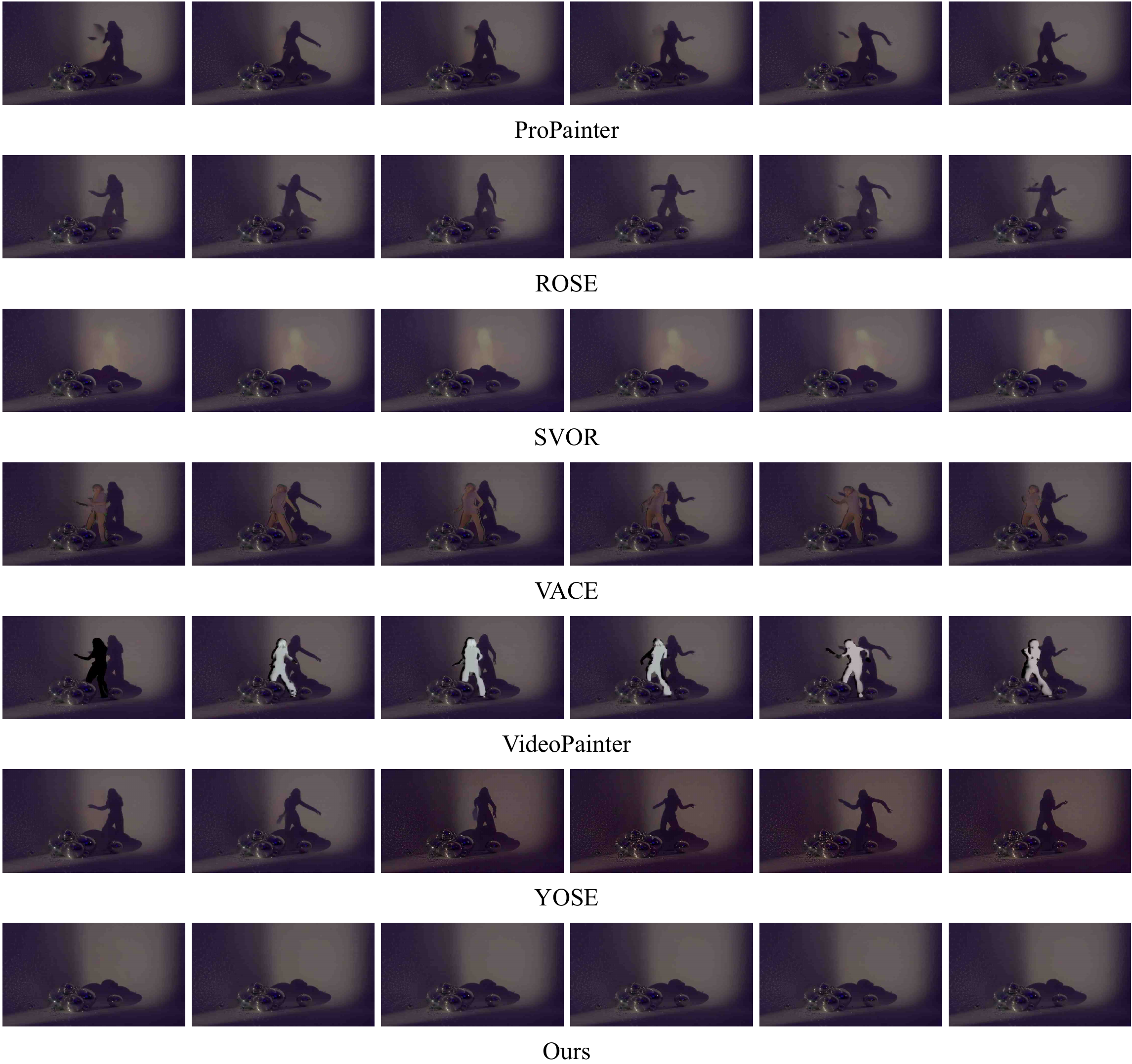}
    \caption{More comparison results on VOR-Wild.(2/2)}
\end{figure*}

\begin{figure*}[t]
    \centering
    \includegraphics[width=0.99\linewidth]{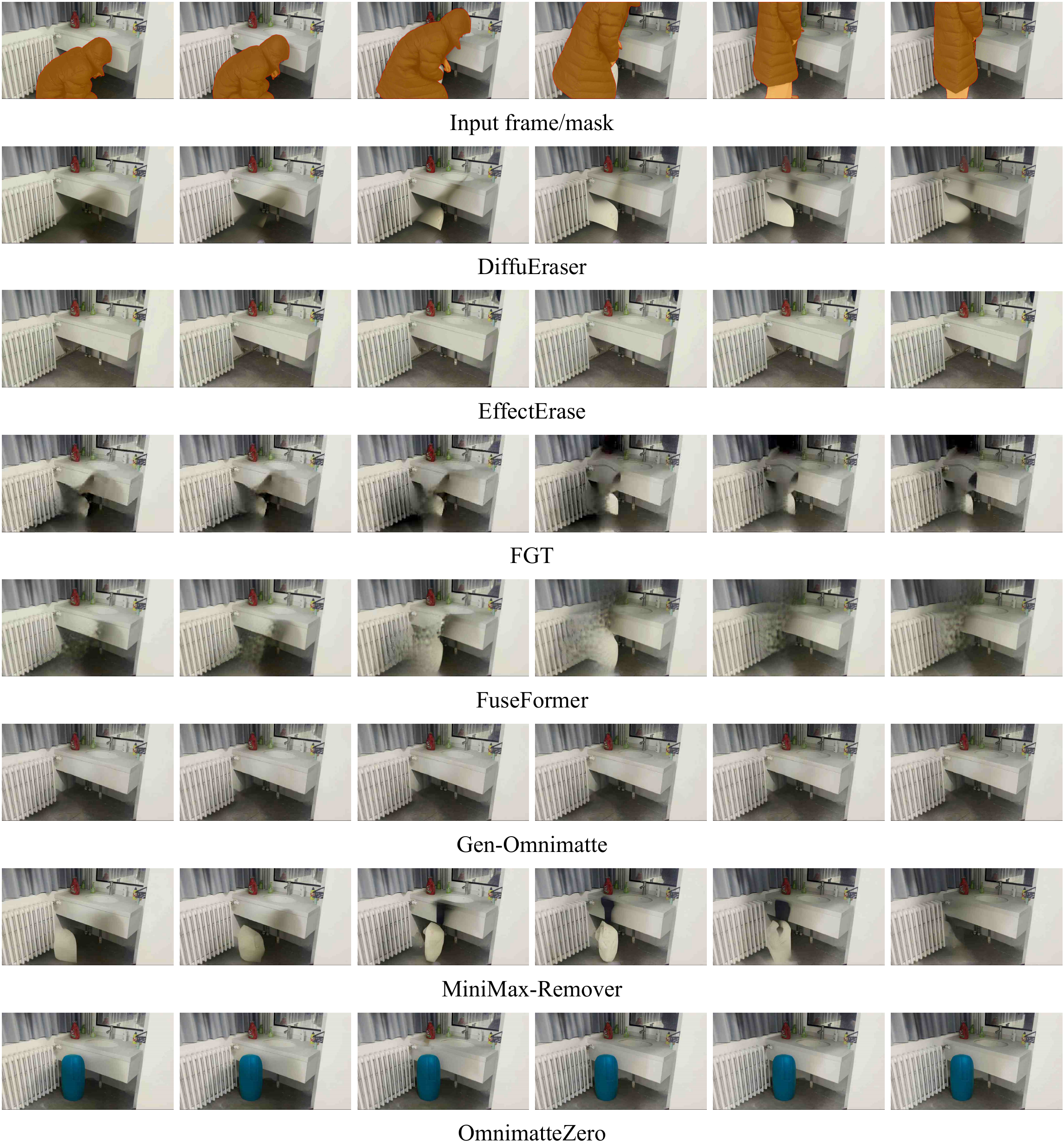}
    \caption{More comparison results on CAMERA-Bench.(1/2)}
\end{figure*}

\begin{figure*}[t]
    \centering
    \includegraphics[width=0.99\linewidth]{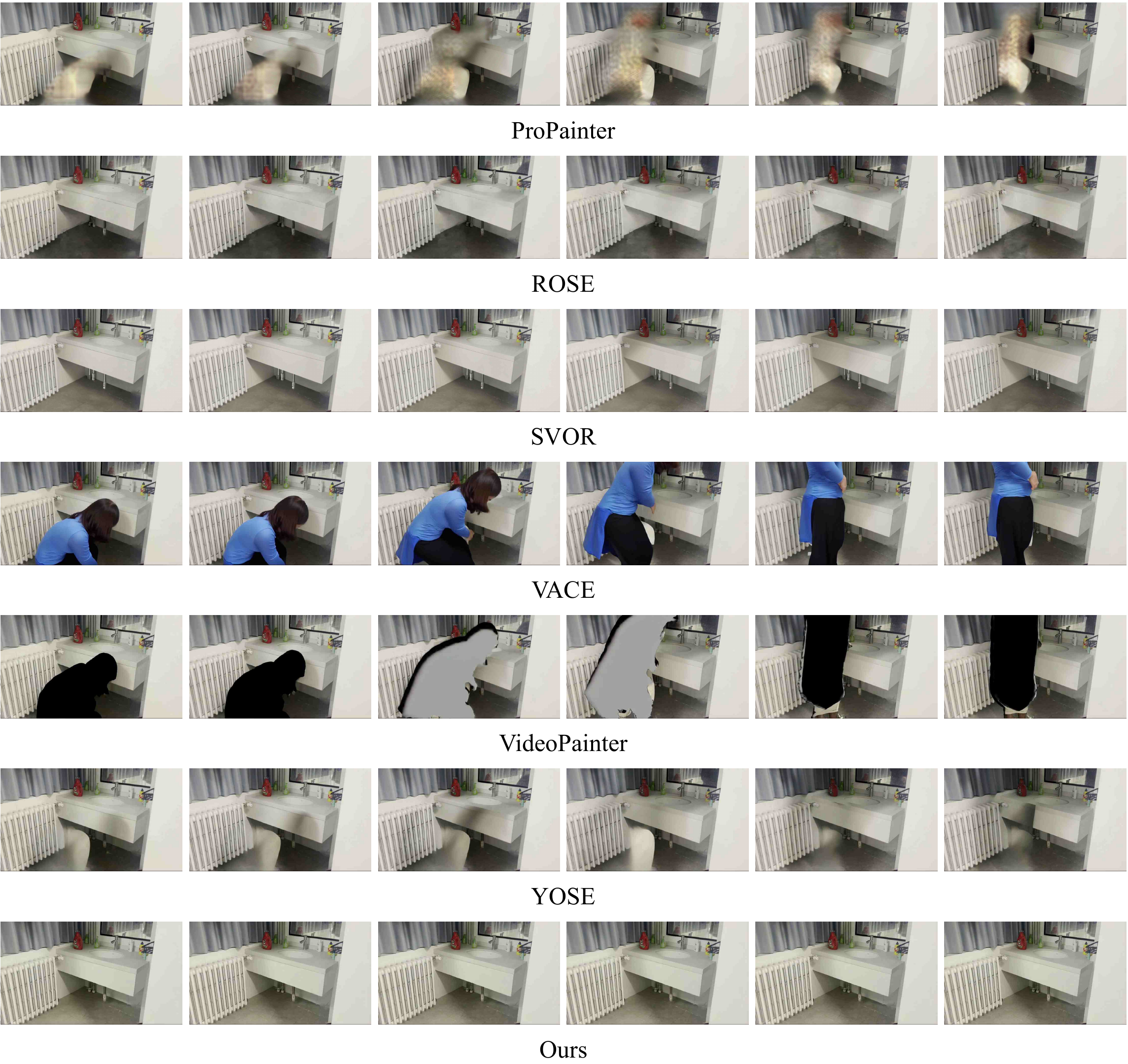}
    \caption{More comparison results on CAMERA-Bench.(2/2)}
\end{figure*}

\begin{figure*}[t]
    \centering
    \includegraphics[width=0.99\linewidth]{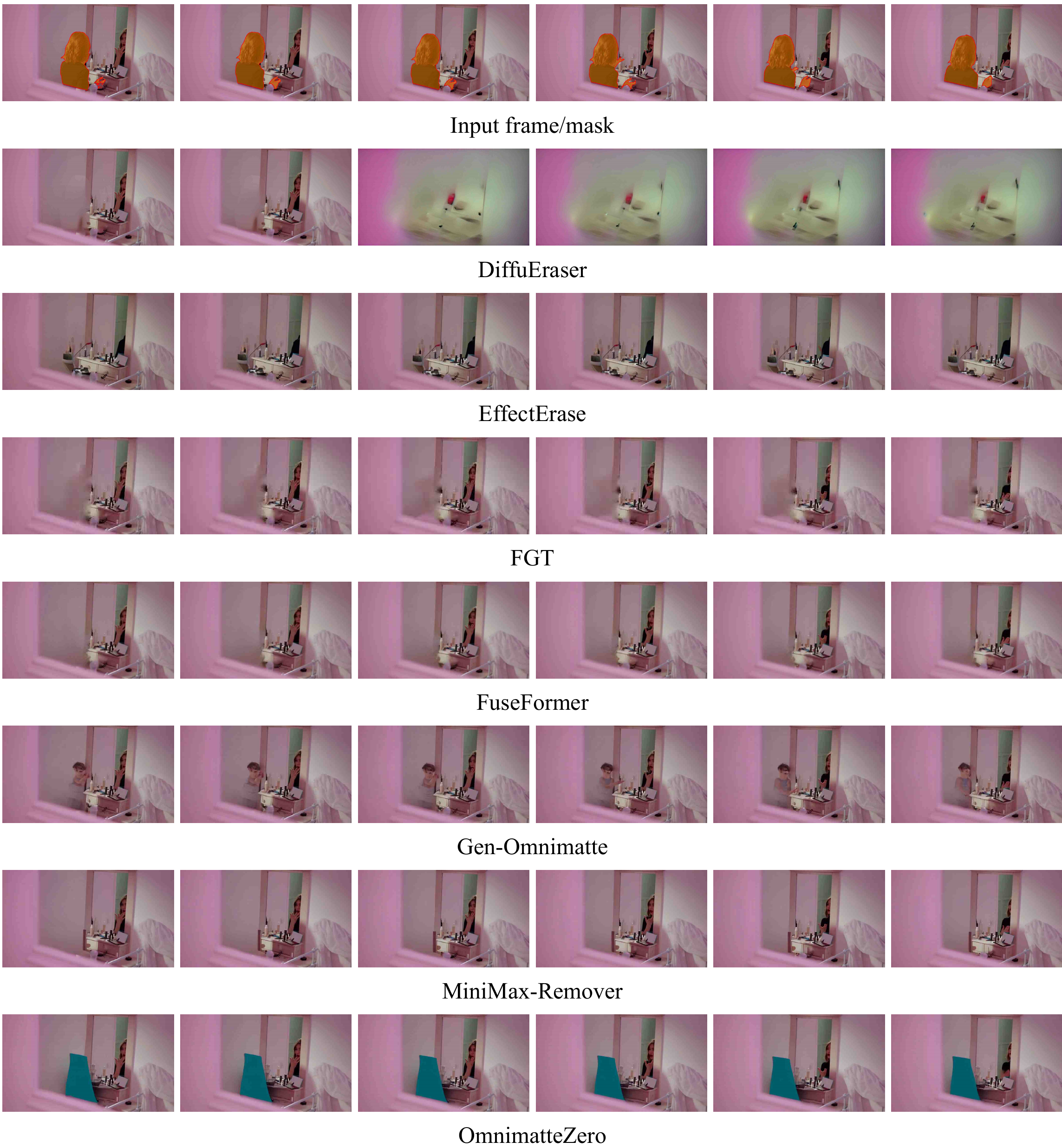}
    \caption{More comparison results on VOR-Wild.(1/2)}
\end{figure*}

\begin{figure*}[t]
    \centering
    \includegraphics[width=0.99\linewidth]{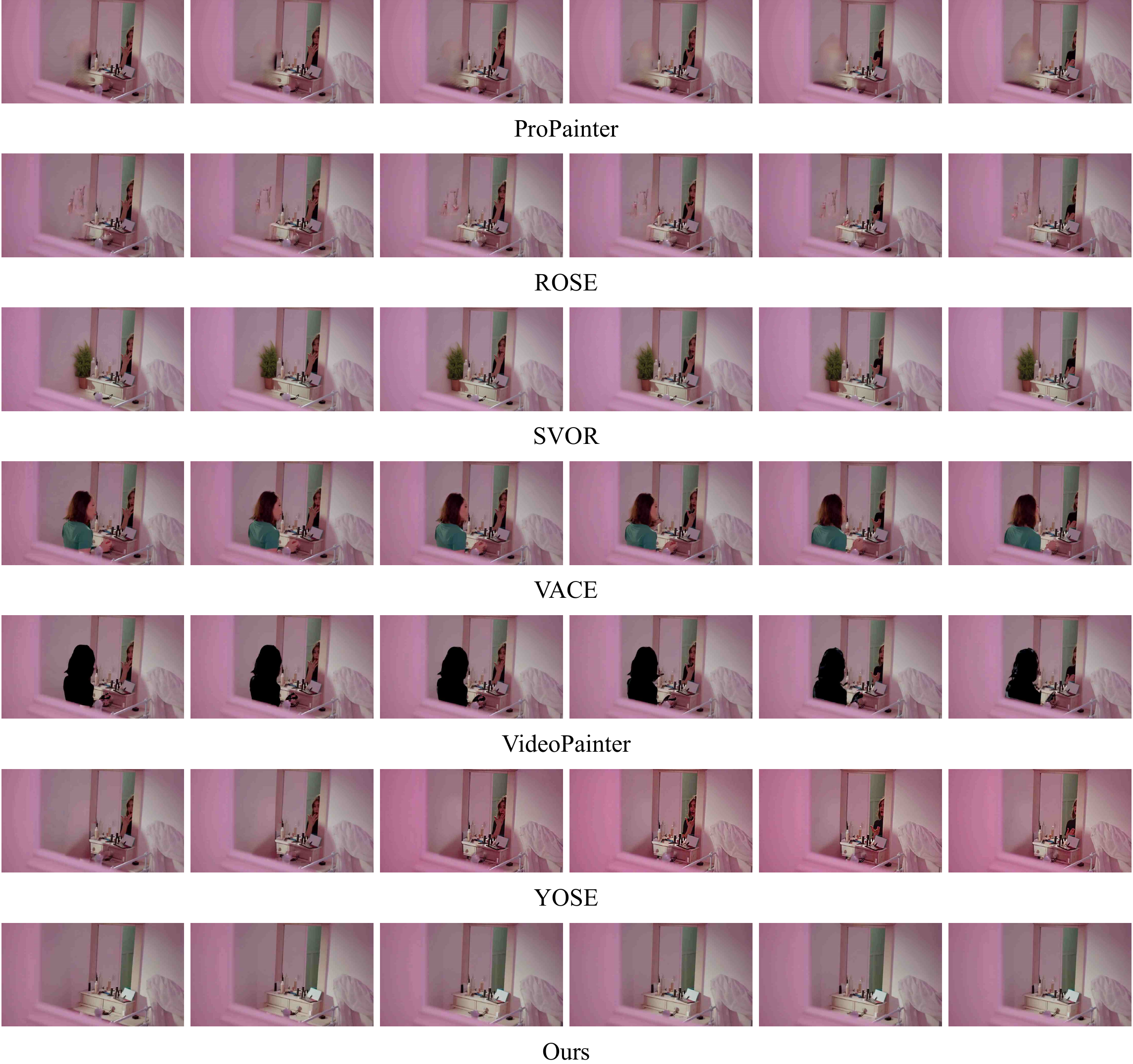}
    \caption{More comparison results on VOR-Wild.(2/2)}
\end{figure*}

\begin{figure*}[t]
    \centering
    \includegraphics[width=0.99\linewidth]{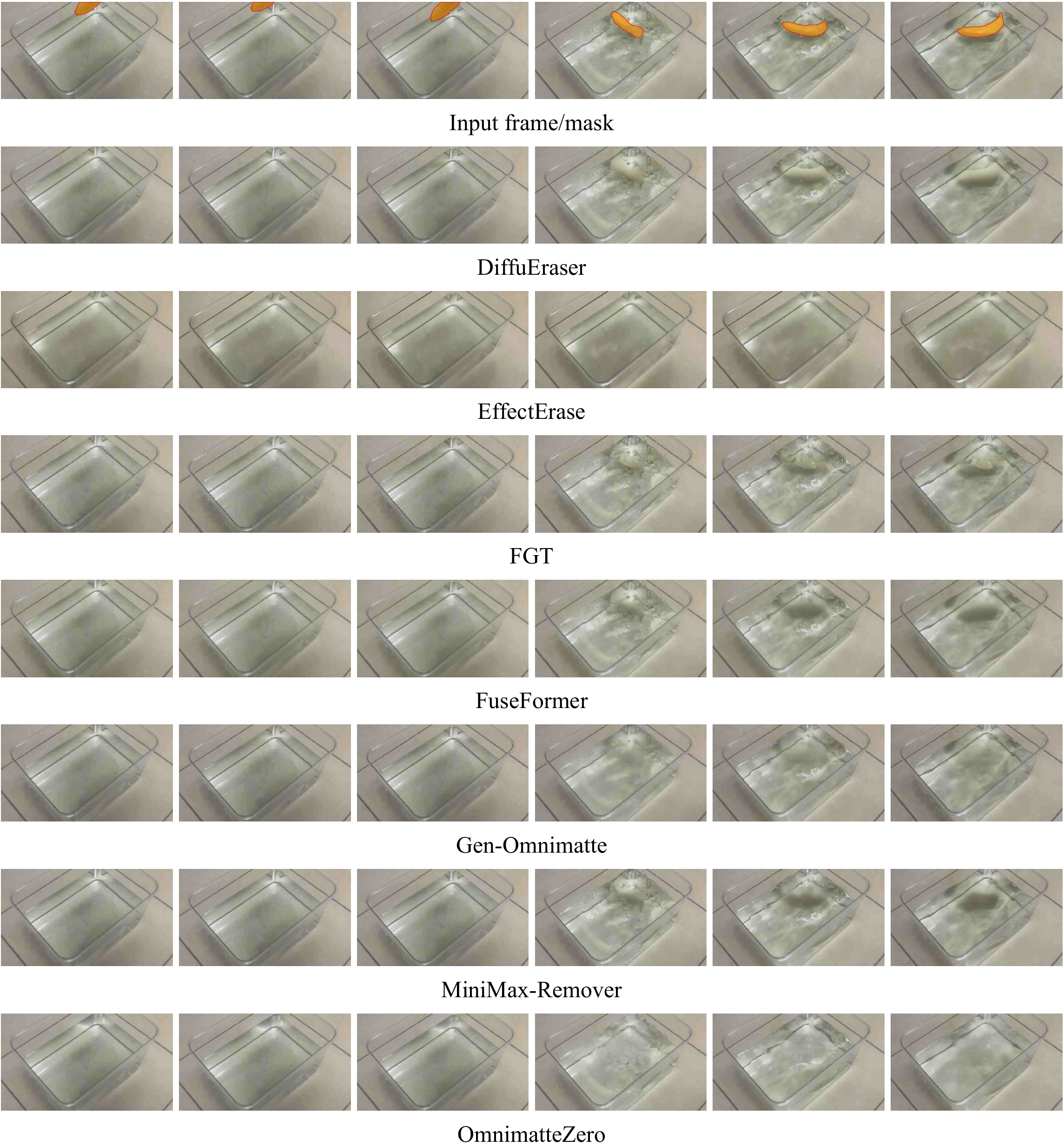}
    \caption{More comparison results on VOR-Eval.(1/2)}
\end{figure*}

\begin{figure*}[t]
    \centering
    \includegraphics[width=0.99\linewidth]{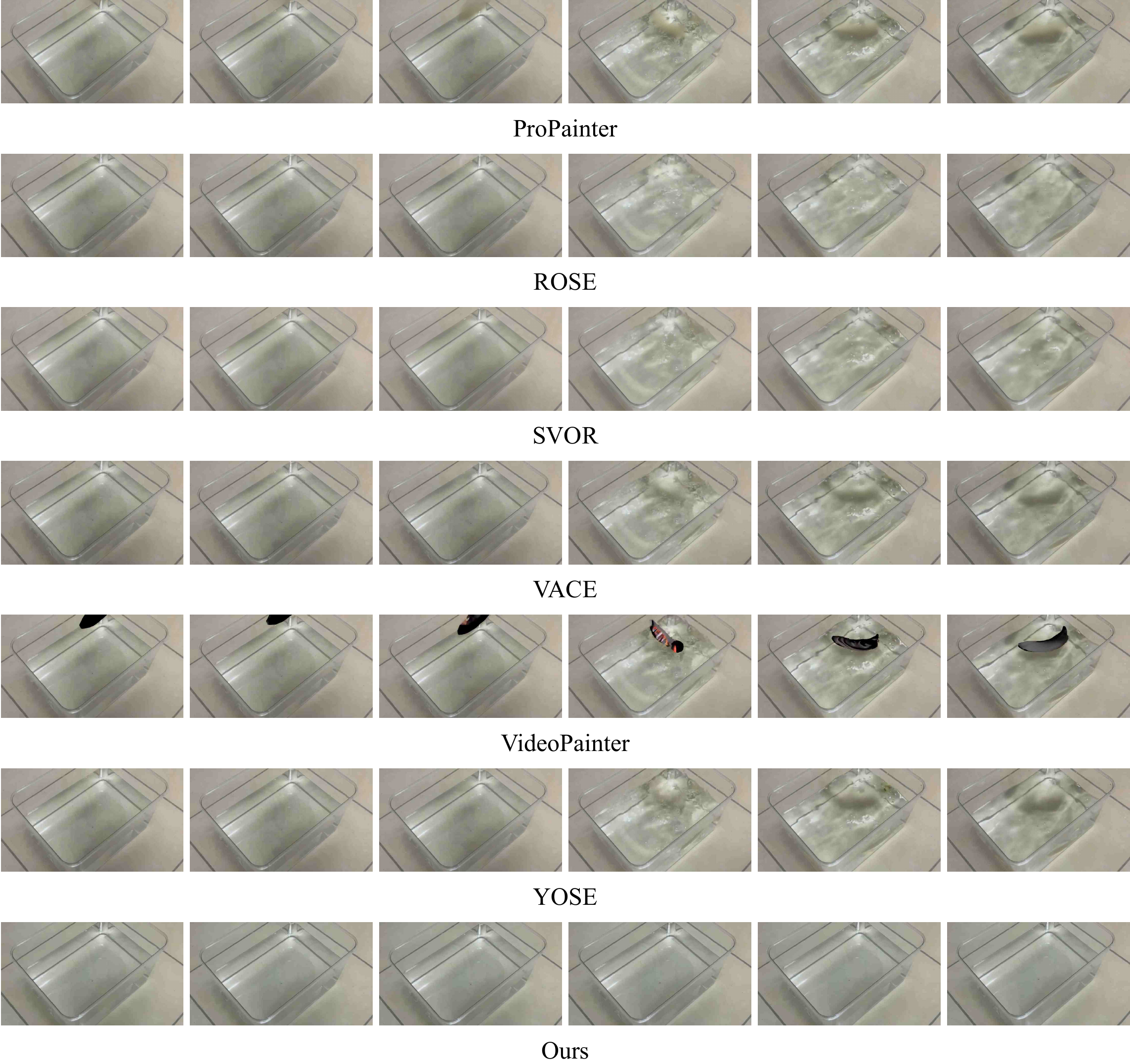}
    \caption{More comparison results on VOR-Eval.(2/2)}
\end{figure*}

\begin{figure*}[t]
    \centering
    \includegraphics[width=0.99\linewidth]{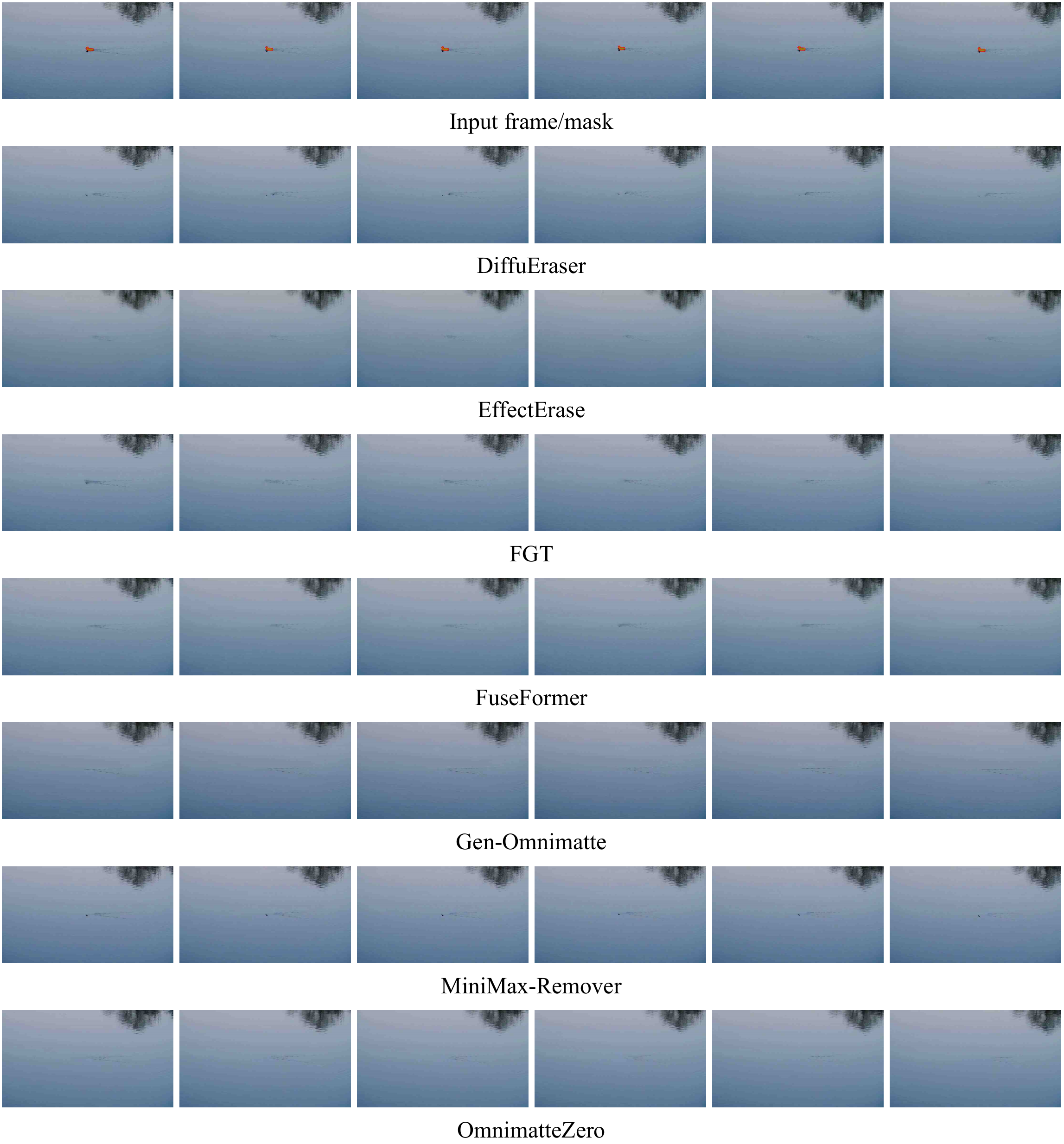}
    \caption{More comparison results on VOR-Wild.(1/2)}
\end{figure*}

\begin{figure*}[t]
    \centering
    \includegraphics[width=0.99\linewidth]{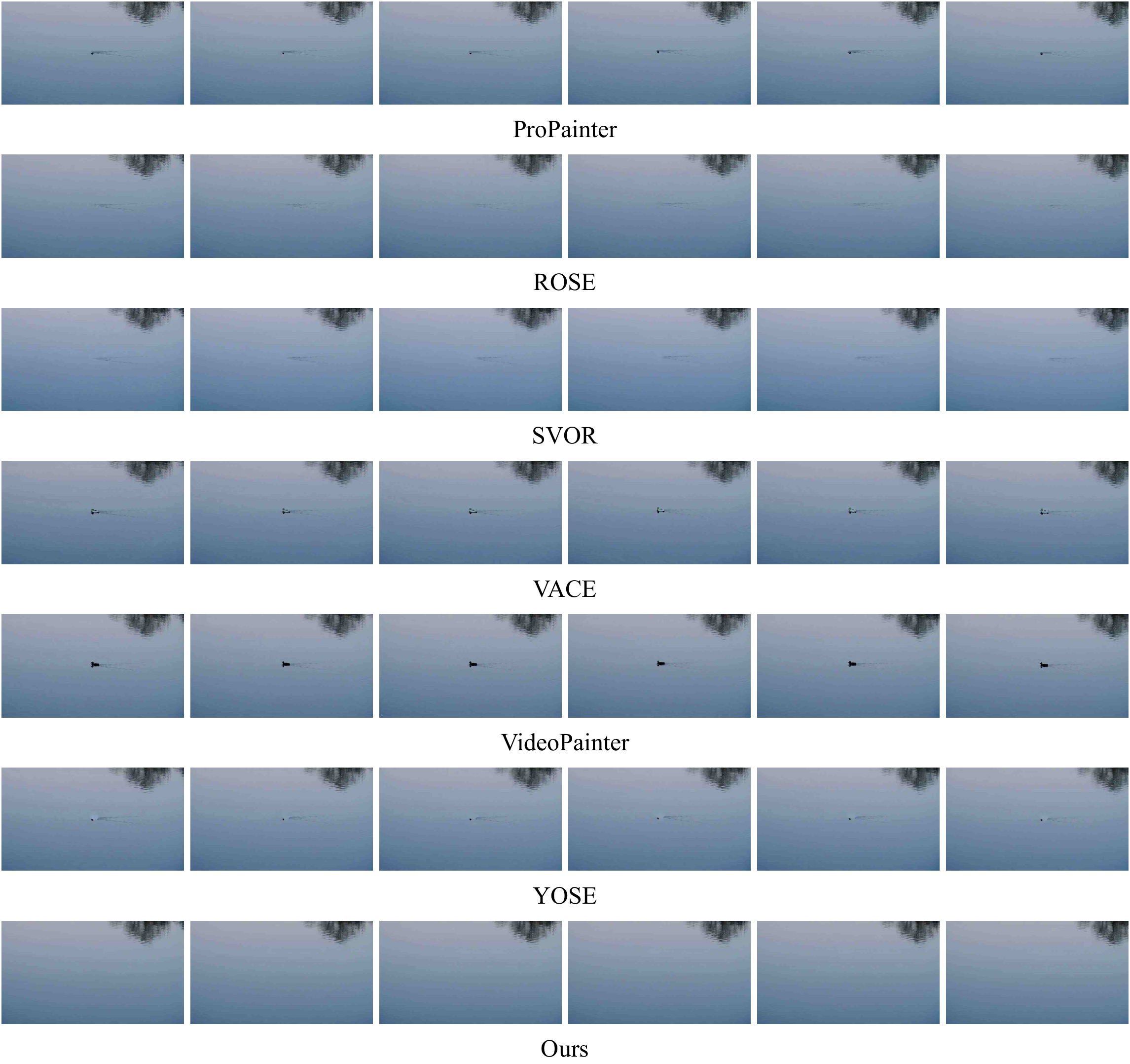}
    \caption{More comparison results on VOR-Wild.(2/2)}
\end{figure*}

\begin{figure*}[t]
    \centering
    \includegraphics[width=0.99\linewidth]{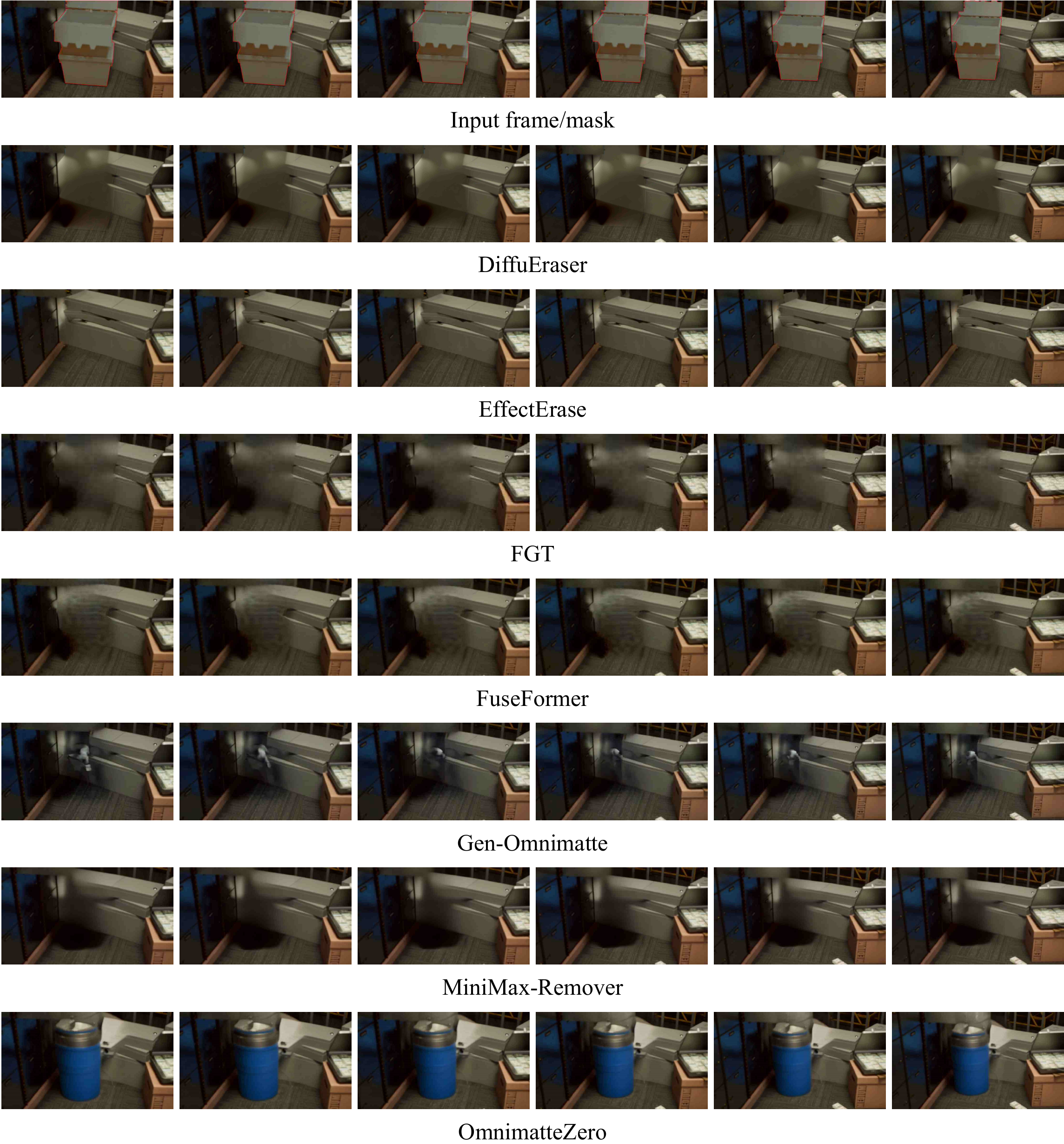}
    \caption{More comparison results on ROSE-Bench.(1/2)}
\end{figure*}

\end{document}